\documentclass[10pt,journal,compsoc]{IEEEtran}
\ifCLASSOPTIONcompsoc
  \usepackage[nocompress]{cite}
\else
  \usepackage{cite}
\fi
\ifCLASSINFOpdf
\else
\fi
\usepackage{epsfig}
\usepackage{helvet}
\usepackage{courier}
\usepackage{floatrow}
\usepackage{multirow}
\newfloatcommand{capbtabbox}{table}[][\FBwidth]
\usepackage{comment}
\usepackage[utf8]{inputenc}
\usepackage[T1]{fontenc}
\usepackage{amsfonts}
\usepackage{units}
\usepackage{microtype}
\usepackage[dvipsnames]{xcolor}
\usepackage{soul}
\usepackage{url}
\usepackage[small]{caption}
\usepackage{graphicx}
\usepackage{amsmath}
\usepackage{booktabs}
\usepackage{epstopdf}
\usepackage{amsthm}
\usepackage{amssymb}
\usepackage{subfigure}
\usepackage{wrapfig}

\usepackage{bbding}
\usepackage{enumitem}
\usepackage{float}
\usepackage{multicol}

\usepackage[ruled,vlined]{algorithm2e}
\usepackage{algpseudocode}
\renewcommand{\algorithmicrequire}{\textbf{Input:}} 
\renewcommand{\algorithmicensure}{\textbf{Output:}}

\usepackage{ragged2e}
\usepackage{colortbl}
\usepackage{hyperref}
\usepackage{soul}

\newcommand\hc{ \rowcolor{teal!20}}

\hypersetup{
	colorlinks=true,
	citecolor=cyan,
}

\usepackage{booktabs}
\usepackage{multirow}
\usepackage{amsmath}
\usepackage{bm}

\usepackage{pifont}
\definecolor{brickred}{rgb}{0.8, 0.25, 0.33}
\definecolor{brickred2}{rgb}{0.25, 0.8, 0.33}
\newcommand{\xm}{\color{brickred2}{\ding{51}}}%
\newcommand{\cm}{\color{brickred}{\ding{55}}}%

\usepackage[switch]{lineno}

\begin{document}

%
\title{\huge Privacy-Preserving SAM Quantization for \\ Efficient Edge Intelligence in Healthcare}
%
%
%
%

\author{Zhikai Li$^*$, Jing Zhang$^*$, and Qingyi Gu 
\IEEEcompsocitemizethanks{
\IEEEcompsocthanksitem $^*$ Equal Contribution
\IEEEcompsocthanksitem This work is supported in part by the National Natural Science Foundation of China under Grant 62276255; in part by the National Science and Technology Major Project under Grant 2022ZD0119402. (Corresponding author: Qingyi Gu)
\IEEEcompsocthanksitem Z. Li, J. Zhang, and Q. Gu are with the Institute of Automation, Chinese Academy of Sciences, Beijing 100190, China, and Z. Li and J. Zhang are also with the School of Artificial Intelligence, University of Chinese Academy of Sciences, Beijing 100049, China. (e-mail: lizhikai2020@ia.ac.cn; zhangjing2024@ia.ac.cn; qingyi.gu@ia.ac.cn)
}
\thanks{Our code will be publicly available.}}

\markboth{}%
{Shell \MakeLowercase{\textit{et al.}}: Bare Demo of IEEEtran.cls for Computer Society Journals}
%



\IEEEtitleabstractindextext{%
\begin{abstract}
\justifying
The disparity in healthcare personnel expertise and medical resources across different regions of the world is a pressing social issue. Artificial intelligence technology offers new opportunities to alleviate this issue by empowering diagnostic and treatment capabilities in underdeveloped areas. Segment Anything Model (SAM), which excels in intelligent image segmentation, has demonstrated exceptional performance in medical monitoring and assisted diagnosis. Unfortunately, the huge computational and storage overhead of SAM poses significant challenges for deployment on resource-limited edge devices, especially in underdeveloped regions with limited equipment and computing power.
Quantization is an effective solution for model compression; however, traditional methods rely heavily on original data for calibration, which raises widespread concerns about medical data privacy and security.
In this paper, we propose a data-free quantization framework for SAM, called DFQ-SAM, which learns and calibrates quantization parameters without any original data, thus effectively preserving data privacy during model compression. Specifically, we propose pseudo-positive label evolution for segmentation, combined with patch similarity, to fully leverage the semantic and distribution priors in pre-trained models, which facilitates high-quality data synthesis as a substitute for real data.
Furthermore, we introduce scale reparameterization to ensure the accuracy of low-bit quantization. 
We perform extensive segmentation experiments on datasets of various modalities such as CT and MRI, and DFQ-SAM consistently provides significant performance on low-bit quantization, e.g., 4-bit quantization results in only a 2.01\% accuracy decrease in abdominal organ segmentation on AbdomenCT1K dataset.
DFQ-SAM decouples the model deployment from real data, eliminating the need for data transfer in cloud-edge collaboration, thereby protecting sensitive data from potential attacks.
By offloading complex medical analysis tasks to local nodes and employing privacy-preserving model compression, it enables secure, fast, and personalized healthcare services at the edge. This enhances system efficiency and optimizes resource allocation, and thus facilitating the pervasive application of artificial intelligence in worldwide healthcare, especially in remote or resource-limited regions.
\end{abstract}}

\maketitle

\IEEEdisplaynontitleabstractindextext

%
\IEEEpeerreviewmaketitle

\IEEEraisesectionheading{\section{Introduction}}
\label{sec:intro}

\IEEEPARstart{T}{he} severe imbalance in healthcare personnel expertise and medical resources worldwide, particularly between developed and underdeveloped regions, has become an urgent social issue\cite{pramesh2022priorities, brito2022global,gelband2016costs}. This disparity prevents patients in underdeveloped regions from receiving timely and effective medical services. For instance, over 70\% of the global cancer burden occurs in low- and middle-income settings, yet most patients with malignancies lack access to the resources and systems available in high-income countries\cite{chalkidou2014evidence}.
Artificial intelligence technology, especially Segment Anything Model (SAM) that excels in intelligent image segmentation\cite{ma2024segment,cheng2023sam,wei2023medsam}, offers unprecedented opportunities to address this issue. SAM demonstrates exceptional performance in medical monitoring and assisted diagnosis, significantly enhancing diagnostic accuracy and treatment outcomes, thereby improving the overall healthcare standards in underdeveloped regions.

\begin{figure}[t]
\centering
\includegraphics[width=0.80\linewidth]{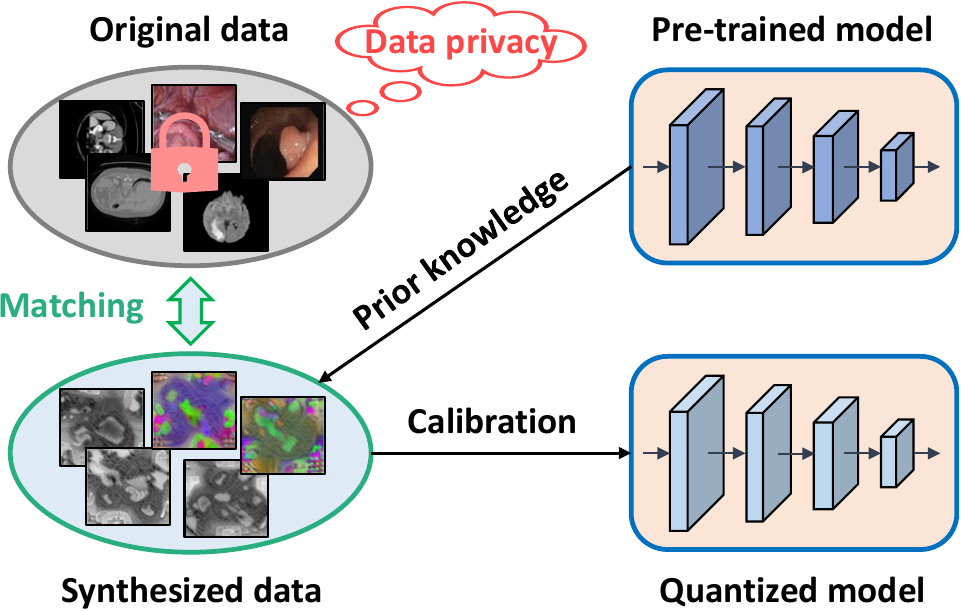}
\caption{Illustration of data-free quantization. It synthesizes data based on information extracted from the pre-trained model, which is then used for quantization calibration. Since no original data is utilized, it effectively protects data privacy and security.}
\label{fig:DFQ}
\end{figure}

\begin{figure*}[t]
\centering
\includegraphics[width=0.96\linewidth]{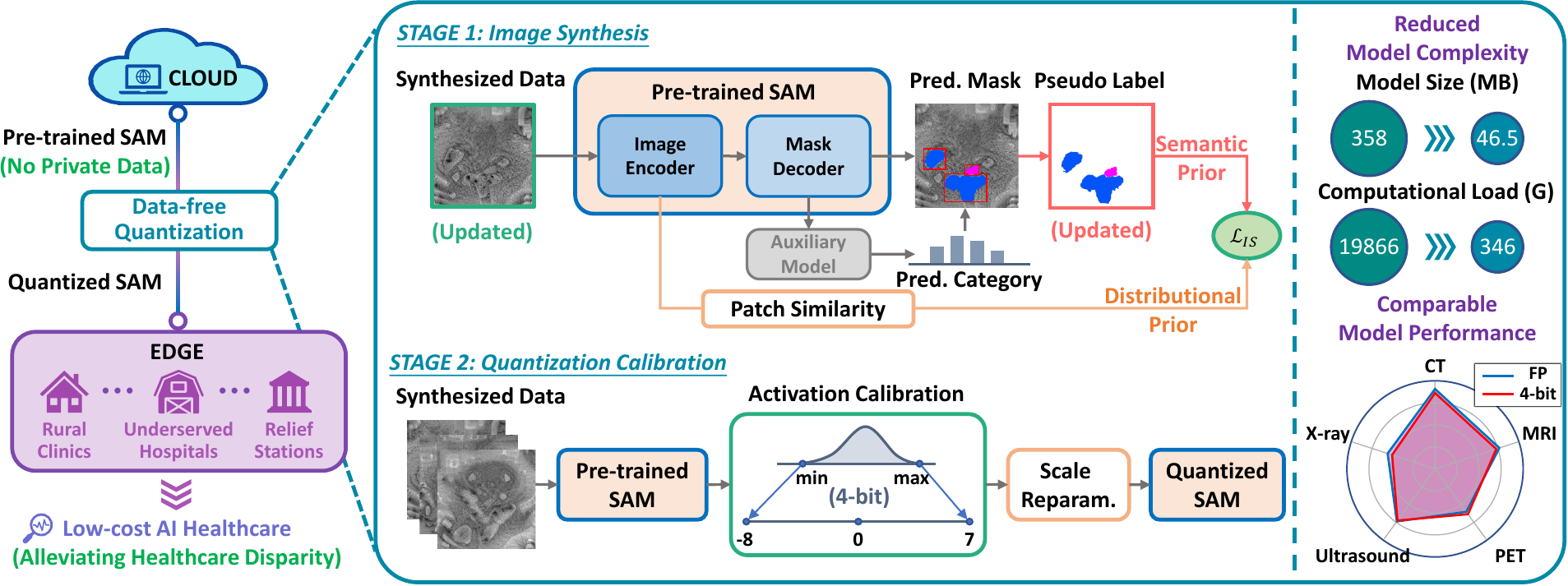}
\caption{Overview of the proposed DFQ-SAM. It retrieves the pre-trained model solely from the cloud, protecting data privacy by avoiding access to any original data. The model is then compressed using data-free quantization for deployment on various edge devices. It enables low-cost and reliable intelligent healthcare at the edge, allowing underdeveloped regions to equally benefit from artificial intelligence technology, thereby helping to alleviate the disparity in healthcare resources.}
\label{fig:overview}
\end{figure*}

Despite the outstanding performance, SAM’s high computational and storage requirements pose significant challenges for its deployment and application\cite{lv2024ptq4sam,zhang2023faster}. This is particularly problematic in underdeveloped regions, where economic constraints limit the capabilities of medical equipment and computing power, which further exacerbates the inequality in healthcare resources, preventing those most in need of technological support from benefiting.
Therefore, to facilitate the downward mobilization of high-quality healthcare resources and ensure convenient access, we are committed to exploring effective model compression techniques that allow SAM to be deployed on low-cost edge devices, serving as dedicated local nodes for each demand source.

Model quantization, which reduces the representation precision of model parameters, is a practical solution for model compression \cite{gholami2022survey,krishnamoorthi2018quantizing}.
Note that considering the diversity in hardware and medical tasks across local nodes, such as different generations of GPUs and CPUs, and various image modalities like CT and MRI, it's essential that each node performs quantization locally with settings tailored to its specific conditions, rather than relying on a unified approach in the cloud.
However, traditional quantization methods require access to the original training data from the cloud to identify parameter distributions for quantization calibration\cite{wang2019haq,esser2019learned,li2024htq}. This dependency on original data poses significant risks to data privacy and security, especially in the data-sensitive medical field\cite{cai2020zeroq,zhong2022intraq,xu2020generative}.
To this end, data-free quantization (DFQ), a.k.a. zero-shot quantization, is increasingly gaining attention\cite{zhang2021diversifying,li2022dual,fan2024data}. Its goal is to utilize the prior information of the pre-trained model to synthesize data in reverse and then use them for calibration, as shown in Fig. \ref{fig:DFQ}. Despite some progress, previous DFQ efforts focus on convolutional neural networks (CNNs) for classification or detection tasks, with little exploration of SAM for segmentation tasks, leaving it an open issue.

\begin{figure*}[t]
\centering
\includegraphics[width=0.85\linewidth]{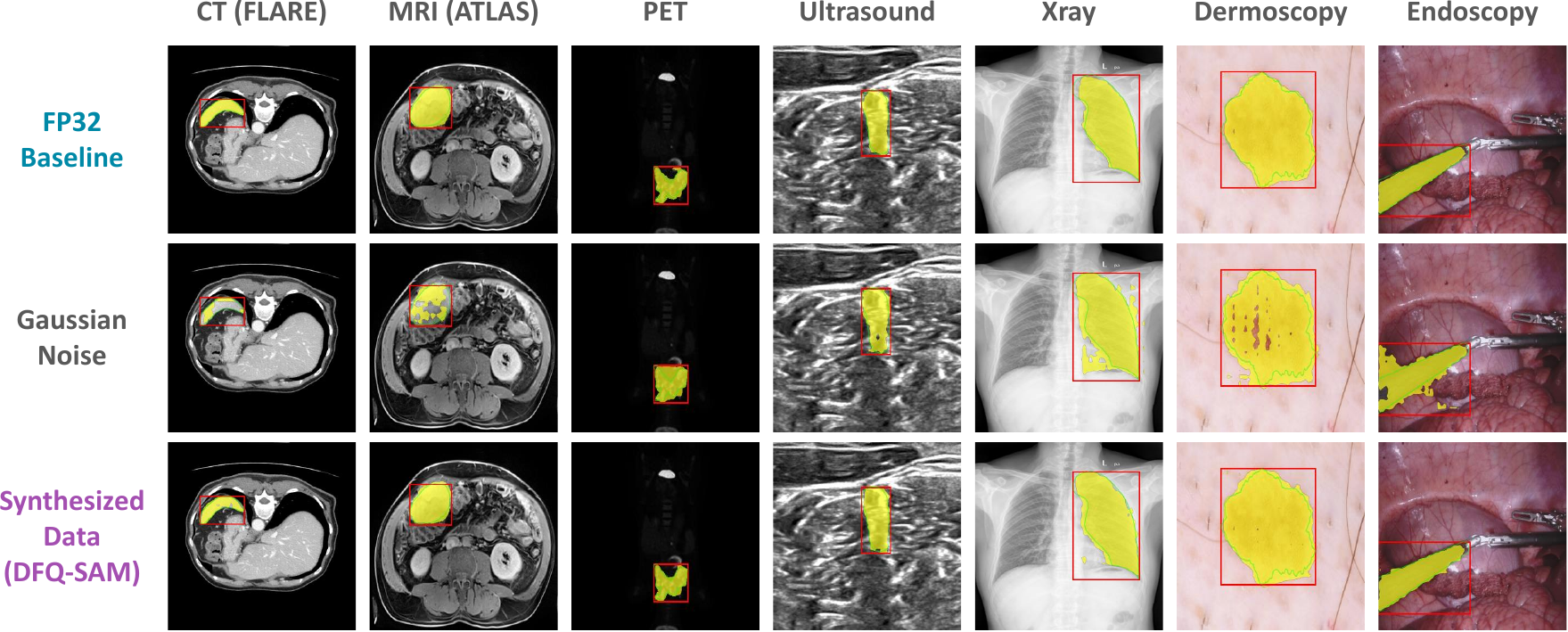}
\caption{Visualization of segmentation results on different datasets (256$\times$256 pixels). The proposed DFQ-SAM, which utilizes synthesized data for 4-bit quantization, consistently performs comparable to full-precision baseline. See Fig. \ref{fig:app_result_vis} for more samples.}
\label{fig:result_vis}
\end{figure*}

For image segmentation with SAM, the dual complexity of its model structure and task presents greater challenges in designing the DFQ algorithm. In terms of model structure, the image encoder in SAM utilizes a Transformer architecture\cite{dosovitskiy2020image}. Most previous works have been limited to CNNs, relying on BatchNorm statistics, which are not applicable to Transformers. Additionally, the DFQ algorithm requires pseudo-labels to guide data synthesis, but segmentation task labels must account for mask size, category, and relative position, making it challenging to specify them appropriately.
To address the above issues, we propose DFQ-SAM, an accurate and practical data-free quantization framework for SAM. Specifically, we propose an evolutionary strategy for pseudo-positive labels, which iteratively samples model outputs to reverse-update the preset labels. This approach ensures that the labels are fully aligned with the response preferences of the pre-trained model, thereby guaranteeing their rationality and effectiveness. We also leverage patch similarity to further exploit the prior knowledge within the Transformer model, enhancing the quality of synthesized data.
Moreover, to achieve accurate quantization calibration, we introduce the scale reparameterization technique, which significantly improves the performance of low-bit quantization.
DFQ-SAM exhibits strong generalization across various image modalities, and the experimental results show that it consistently achieves impressive performance on datasets of multiple modalities such as CT and MRI. For instance, when performing 4-bit quantization, DFQ-SAM achieves an 8$\times$ reduction in model size and a 64$\times$ reduction in computational complexity (i.e., BOPs), while only incurring a 2.01\% drop in accuracy on AbdomenCT1K dataset.

\noindent\textbf{Discussion}
Note that this work is an extension of our previous works, PSAQ-ViT\cite{li2022patch,li2023psaq} and RepQ-ViT\cite{li2023repq}, which lay a strong theoretical foundation and provide early validation. PSAQ-ViT introduces patch similarity, making the first successful attempt at data-free quantization for Transformers, and RepQ-ViT is a powerful baseline of low-bit quantization calibration in the community. Building on these achievements, DFQ-SAM introduces a novel segmentation label evolution strategy tailored for SAM, resulting in a promising and practical data-free quantization framework. To the best of our knowledge, it is the first data-free quantization approach for SAM.
DFQ-SAM has the potential to accelerate the widespread adoption of SAM-based intelligent healthcare through low-cost, multi-node distributed edge deployment, as shown in Fig. \ref{fig:overview}. In this system, a large central model is maintained in the cloud, where it is continuously updated and upgraded, and periodically distributed to edge nodes. At the edge, the model is quantized according to the hardware specifications and used to deliver medical services. Thanks to data-free quantization, only the pre-trained model needs to be transmitted between the cloud and the edge, eliminating the need for any training data. This not only greatly reduces bandwidth requirements but, more importantly, ensures robust protection of data privacy and security. 
Therefore, DFQ-SAM effectively addresses data privacy concerns in the compression and deployment of SAM, facilitating its widespread and secure application in real-world scenarios. In the context of the current imbalance in healthcare resource distribution, this technology helps underdeveloped regions equally benefit from intelligent healthcare, making a significant impact on improving medical conditions in these areas.

\section{Related Works}

\noindent\textbf{Medical SAM}
Medical image segmentation, which works on separating different tissues, organs, or pathological areas within complex medical images, can provide clear diagnostic insights and treatment guidance, and thus plays a crucial role in clinical practice\cite{patil2013medical,wang2021annotation}.
In recent years, a series of notable deep learning-based models, such as UNet++\cite{zhou2018unet++} and TransUNet\cite{chen2021transunet}, have emerged. However, these models are typically designed for specific image modalities or anatomical regions and lack generalizability. In contrast, SAM stands out due to its versatility, zero-shot transfer capabilities, and interactive features\cite{kirillov2023segment}, resulting in a significant improvement in segmentation performance. To enhance the adaptability of SAM to medical images, MedSAM\cite{ma2024segment} and SAM-Med 2D\cite{cheng2023sam} collect large-scale medical image datasets to fine-tune the model, significantly improving SAM's performance on medical image segmentation. There are also works that extend them to applications with higher image resolutions\cite{wei2023medsam} and 3D scenarios\cite{Wang2023SAMMed3D}.

However, SAM's high model complexity makes it challenging to deploy and execute on resource-constrained edge devices, particularly in the medical field where high concurrency and real-time processing are essential. FastSAM\cite{zhao2023fast} and MobileSAM\cite{zhang2023faster} optimize model efficiency from an architectural design perspective, but there is still a large gap with practical implementation.
Consequently, model compression techniques for SAM are critically desired to enable real-time intelligent medical analysis at the edge.

\noindent\textbf{Data-free Quantization}
Quantization, which uses reduced numerical precision to represent parameters, is an effective approach for model compression\cite{gholami2022survey,li2023vit}. In the quantization process, parameter calibration relies on the numerical distribution, which traditionally requires the input of original training data to obtain this distribution\cite{esser2019learned,li2024repquant}. This approach raises data privacy and security concerns, particularly in the medical field where access to data is typically restricted, rendering traditional methods impractical\cite{cai2020zeroq,zhong2022intraq}. To address these challenges, various efforts have begun to explore the emerging field of data-free quantization\cite{xu2020generative,zhang2021diversifying,fan2024data}. It focuses on leveraging the prior knowledge embedded in pre-trained models to synthesize data for calibration, and the quality of synthesized data is crucial to quantization performance\cite{li2023psaq}.

The quality of synthesized data depends on its consistency with the original training data in both distribution and semantics. Earlier approaches achieve distribution consistency through BatchNorm statistics\cite{cai2020zeroq,zhang2021diversifying} and promote semantic consistency using preset pseudo-labels\cite{chawla2021data,zhong2022intraq}. However, they suffer from great limitations on model structures and tasks. On one hand, there is no BatchNorm layer in Transformers, resulting in unavailable statistical information; on the other hand, the presetting of labels for segmentation tasks is unexplored. Therefore, how to realize data-free quantization of SAM for image segmentation remains an open issue.
In this paper, we first employ patch similarity as a substitute for BatchNorm statistics to capture the prior distribution in Transformers, as proposed in our preliminary work PSAQ-ViT\cite{li2022patch,li2023psaq}. Building on it, we propose the pseudo-positive label evolution strategy to generate and iteratively update the segmentation labels, thus achieving accurate data-free quantization of SAM.

\section{Methods}

\begin{figure*}[t]
\centering
\includegraphics[width=0.9\linewidth]{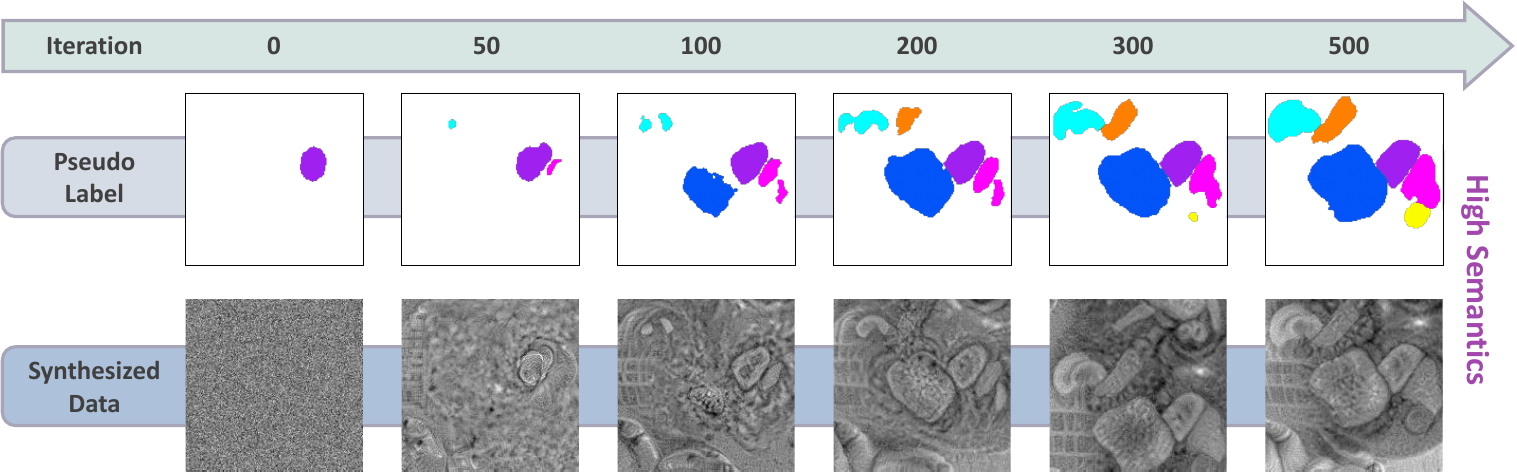}
\caption{Illustration of pseudo-label evolution and synthesized image updating (256$\times$256 pixels). Label evolution and image updating are conducted alternately: The label continuously discovers new regions based on the image outputs, which in turn further drives the image optimization. Ultimately, both converge at a high semantic level. See Fig. \ref{fig:app_label_vis} for more samples.}
\label{fig:label_vis}
\end{figure*}

\subsection{Preliminaries: Quantization}
Model quantization discretizes the full-precision parameters in the pre-trained model and represents them as integer values.
In the following, we introduce the discretization process through the uniform quantizer, which is the simplest and most commonly used quantizer:
\begin{equation}
\begin{aligned}
  Quant&: \bm{x}^{\mathbb{Z}} = \text{clip}\left(\left\lfloor \frac{\bm{x}}{s} \right\rceil+z, 0, 2^b-1 \right), \\
  DeQuant&: \hat{\bm{x}} = s\left(\bm{x}^{\mathbb{Z}}-z\right) \approx \bm{x},
\end{aligned}
\end{equation}
where $\bm{x}\in \mathbb{R}$ and $\bm{x}^{\mathbb{Z}}\in \mathbb{Z}$ represent the floating-point and quantized values, respectively. The operator $\left\lfloor\cdot\right\rceil$ refers to the rounding function, and $b \in \mathbb{N}$ is the bit-width used for quantization. The de-quantized value $\hat{\bm{x}}$ serves as an approximation to $\bm{x}$.
Notably, $s\in \mathbb{R}^+$ is the quantization scale, and $z \in \mathbb{Z}$ is the zero-point. Both of them are essential parameters for quantization and they are determined by the numerical distribution of $\bm{x}$ as follows:
\begin{equation}
\label{eq:sz}
s = \frac{\max(\bm{x})-\min(\bm{x})}{2^b-1}, \quad z = \left\lfloor-\frac{\min(\bm{x})}{s} \right\rceil.
\end{equation}

Formally, Eq. \ref{eq:sz} is referred to as the quantization calibration procedure. It is evident that accurately capturing the parameter distributions to obtain precise values for $s$ and $z$ is crucial. Traditionally, acquiring the distribution of activation values requires inputting the original training data to extract feature maps. However, this approach raises privacy and security concerns, particularly in sensitive fields like healthcare. Thus, in this paper, we are dedicated to exploring how to perform quantization calibration without accessing the original data.

\subsection{Data Synthesis from Pre-trained SAM}
The quality of synthesized data from pre-trained models is measured in two ways: semantic matching and distributional matching. We design SAM-specific methods to facilitate both matching, as described in detail below.

\noindent\textbf{Semantic Matching: Pseudo-positive Label Evolution}
To enhance the semantic information of synthesized images, we need to predefine the labels as a guide and utilize the pre-trained model as a discriminator to optimize the images by facilitating the alignment of the model outputs with the labels. For instance, in the classification task, if the category label "dog" is preset, the image is optimized so that when it is input into the model, the output predicts "dog" with progressively higher confidence, thereby infusing the image with clear and distinct "dog" semantics. As a result, the semantics of the final synthesized images are strongly correlated with the predefined labels. However, creating labels with reasonable semantics for segmentation tasks is challenging, as it requires consideration of the mask's size, category, and relative position. 

Therefore, instead of direct manual predefinition, we propose the pseudo-positive label evolution strategy, which continuously samples the model's output in iterations and progressively superimposes high-confidence (i.e., pseudo-positive) predicted masks into the labels. This capitalizes on the output preferences of the pre-trained model, allowing the pseudo-labels to be fully aligned with the semantics learned by the model.
Interestingly, the concept of label evolution is also utilized in DIODE \cite{chawla2021data} for object detection. However, in this work, the segmentation task and the zero-shot capabilities of SAM create a stark contrast, posing greater challenges for mask sampling: (i) Masks are pixel-level, with irregular shapes and connectivity; (ii) SAM's outputs do not include category information.

To address the above challenges, we measure confidence by calculating the average category score within the mask region, with the category scores obtained from an auxiliary model (e.g., TransUNet \cite{chen2021transunet}). Formally, the computation process is as follows:
\begin{equation}
\begin{aligned}
    M_{c^*}^{(t)} = \arg \max \left(\frac{1}{\left|M_c\right|} \sum_{(h, w) \in M_c} S_{h, w, c}\right),
    \label{eq:M_c}
\end{aligned}
\end{equation}
where $M_c$ is the predicted mask of category $c$, with region provided by SAM and category provided by the auxiliary model, and $|M_c|$ is the number of pixels in the mask.
$S_{h, w, c}$ is the confidence score that the pixel ($h$,$w$) belongs to the category $c$.  $M_{c^*}^{(t)}$ is the pseudo-positive response obtained at the $t$-th iteration.
To ensure the saliency of sampling, we continuously filter them in real time according to confidence scores and connected component sizes during the iteration process. Eventually, we stack the filtered masks and obtain the pseudo-labels as follows:
\begin{equation}
    \mathcal{M}^{GT} = \{M_{c^*}^{(t)} \mid \max_{(h,w)\in M_{c^*}^{(t)}}(S_{h, w, c^*})>\varepsilon_1, |M_{c^*}^{(t)}|>\varepsilon_2\},
    \label{eq:m_gt}
\end{equation}
where $\varepsilon_1$ and $\varepsilon_2$ are the filter thresholds for confidence and size of the mask, respectively. Given the pseudo-labels, we define \emph{Semantic Matching Loss} to enhance the semantic content of synthesized images as follows:
\begin{equation}
    \mathcal{L}_{SM} = \mathcal{L}_{Mask} + \alpha \cdot \mathcal{L}_{Class},
\end{equation}
\begin{equation}
    \begin{aligned}
        \text{where} \quad \mathcal{L}_{Mask} &= 1 - \frac{|\mathcal{M}^{Pred} \cap \mathcal{M}^{GT}|}{|\mathcal{M}^{Pred} \cup \mathcal{M}^{GT}|},  \\
        \mathcal{L}_{Class} &= - \frac{1}{|\mathcal{M}^{GT}|} \sum_{M_{c^*} \in \mathcal{M}^{GT}} \sum_{(h,w)\in M_{c^*}} \log(S_{h, w, c^*}).
    \end{aligned}
    \label{eq:l_sm}
\end{equation}

Here, $\mathcal{L}_{Mask}$ is used to promote IoU matching of mask regions, $\mathcal{L}_{Class}$ is conditional cross entropy to promote category matching within the masks, and $\alpha$ is the balance coefficient. Note that label evolution and image updating occur alternately: The synthesized images provide response preferences for label evolution while the resulting labels guide image updating, and ultimately both converge at a high semantic level.

Fig. \ref{fig:label_vis} illustrates the process of pseudo-label evolution and synthesized image updating. The label is initialized by a mask with random position, size, and category, and the image is initialized by Gaussian noise. As the iterations proceed, the label evolves: in the 50th iteration, a liver region (red) appears near the right kidney (purple); in the 200th iteration, a pancreas (orange) region appears near the stomach (dark blue). Also, the left kidney (bright blue) and the right kidney (purple) are distributed symmetrically on both sides of the stomach (dark blue). These present spatial relationships aligned with the real labels. Moreover, the shapes and sizes of the masks also gradually converge to the real labels. For instance, the kidneys (purple and bright blue) appear elliptical or auricular, the pancreas (orange) is in the form of a thin irregular strip, and the stomach (dark blue) is relatively large. As a result, both the label and the image gradually learn abundant semantic information, resulting in a mutually beneficial situation.

\noindent\textbf{Distributional Matching: Patch Similarity Metric}
Previous methods are designed for CNNs, and they rely on BatchNorm statistics to obtain the distribution of the original data. However, Transformers employs LayerNorm, which does not store any previous information. To this end, our preliminary work PSAQ-ViT \cite{li2022patch} summarizes the general difference between the two by examining the self-attention module's response in Transformers when the inputs are real images and Gaussian noise, respectively. The details are in Appendix \ref{app:pasq}. Building on this insight, it designs a relative value metric, the diversity of patch similarity, to quantify this difference and optimizes it during image synthesis to narrow the gap, allowing the synthesized image to gradually approach the distribution of real images from the initial Gaussian noise.

Specifically, we first hook the output from the attention module, denoted as $O_l$. Here, $l\in \{1,\cdots,L\}$, and $L$ is the number of layers.
Then, we normalize $O_l$ in its patch dimension to ensure a unified range of relative value metrics, which is achieved by computing the cosine similarity as follows:
\begin{equation}
    \bm{\Gamma}_l\left(u_i, u_j\right)=\left\{\left.\frac{u_i \cdot u_j}{\left\|u_i\right\|\left\|u_j\right\|} \right\rvert\, i, j \in\{1,2, \ldots, N\}\right\},
    \label{eq:gamma}
\end{equation}
where $u_i$ and $u_j$ are the $i$-th and $j$-th vectors in the patch dimension of $O_l$, respectively, and $N$ is the number of patches. Here, $\bm{\Gamma}_l$ is referred to as \emph{patch similarity}.

The diversity of patch similarity is measured by the differential entropy, which is calculated as follows:
\begin{equation}
\begin{aligned}
    \mathbb{E}(\bm{\Gamma}_l;\mathcal{G}) = -\int f_h(x) \cdot \log \left[ f_h(x) \right] dx, \\
    \text{where} \quad f_h(x) = \frac{1}{K}\sum_{k=1}^{K}\mathcal{K}_h(x-x_k).
    \label{eq:entropy}
\end{aligned}
\end{equation}

Here, $f_h(x)$ is the continuous probability density function of $\bm{\Gamma}_l$, estimated using kernel density estimation. $\mathcal{K}_h(\cdot)$ represents the kernel (e.g. normal kernel) with the bandwidth $h$, $x_k$ is a training point drawn from $\bm{\Gamma}_l$ and serves as the center of a kernel, and $x$ is the given test point.

Finally, we sum the differential entropy of each layer to capture the diversity of patch similarity across all layers, and the \emph{Distributional Matching Loss} is defined as follows:
\begin{equation}
	\mathcal{L}_{DM} = \sum_{l=1}^{L}\mathbb{E}(\bm{\Gamma}_l;\mathcal{G}).
	\label{eq:l_dm}
\end{equation}

\subsection{Quantization Calibration: Scale Reparameterization}
With the synthesized data, it is also crucial to utilize them for accurate quantization calibration. However, the Transformer structure in SAM poses significant challenges for low-bit quantization due to the extreme parameter distributions of its unique components. In particular, LayerNorm activations exhibit severe inter-channel variations, so that sharing a unified quantization scale across all channels leads to crashing quantization performance.
Therefore, we propose a novel quantization strategy, which first computes a unique quantization scale for each channel, and then converts them into a unified scale through mathematical equivalent transformations. This can significantly improve the quantization accuracy while ensuring hardware adaptability. 

Specifically, we perform an independent quantization calibration for each channel to obtain the quantization scale $\bm{s}$ and zero-point $\bm{z}$. Here, both $\bm{s}$ and $\bm{z}$ are $D$-dimensional vectors, and $D$ is the number of channels in LayerNorm activations. Then, we define the reparameterization procedures $\tilde{s}=\text{E}[\bm{s}]$ and $\tilde{z}=\text{E}[\bm{z}]$, where $\tilde{s}$ and $\tilde{z}$ are the values to be shared by all channels.
To hold the model output constant after reparameterization, we need to adjust the affine factors $\bm{\beta}$ and $\bm{\gamma}$ in LayerNorm and the next layer's weights $\bm{W}^{qkv}$ and $\bm{b}^{qkv}$ as follows:
\begin{align}
\left\{
\begin{aligned}
\widetilde{\bm{\beta}} &= (\bm{\beta}+\bm{s}\odot \bm{r}_2) / \bm{r}_1\\
\widetilde{\bm{\gamma}} &= \bm{\gamma}/ {\bm{r}_1}\\
  \widetilde{\bm{W}}^{qkv}_{:,j} &= \bm{r}_1\odot\bm{W}^{qkv}_{:,j} \\
  \widetilde{\bm{b}}^{qkv}_j &= \bm{b}^{qkv}_j - (\bm{s}\odot \bm{r}_2) \bm{W}^{qkv}_{:,j} 
\end{aligned}
\right.
\label{eq:raparm}
\end{align}
where $\bm{r}_1=\bm{s}/\tilde{\bm{s}}$ and $\bm{r}_2=\bm{z}-\tilde{\bm{z}}$ are the variation factors, $j\in\{1,\cdots,D'\}$, and $D'$ is the number of channels in $\bm{W}^{qkv}$.
Eq. \ref{eq:raparm} can realize equivalent transformations from $\bm{s}$ and $\bm{z}$ to $\tilde{s}$ and $\tilde{z}$ while maintaining the model output the same.
Note that the equivalent transformations are proposed in our preliminary work RepQ-ViT \cite{li2023repq}, and the details of the derivation are in Appendix \ref{app:repq}.

\renewcommand{\algorithmicrequire}{\textbf{Input:}}
\renewcommand{\algorithmicensure}{\textbf{Output:}}
\begin{algorithm}[t]
    \small
    \caption{The DFQ-SAM Pipeline}
    \label{alg:1}
    \KwIn{Pre-trained SAM $\mathcal{P}$.}
    \KwOut{Synthesized images $I_S$,  Quantized SAM $\mathcal{Q}$.}
    Initialize $I_S$ from Gaussian distribution $\mathcal{N}(0,1)$\;
    {\color{teal}\# \textbf{Stage 1: Image Synthesis}}\\
    \For{$t=1,2,\cdots$}
    {
    Input $I_S$ to $\mathcal{P}$ and forward propagate $\mathcal{P}(I_S)$\;
    {\color{teal}\# Semantic Matching}\\
    Collect maximum response preference $M_{c^*}^{(t)}$ by Eq. \ref{eq:M_c}\;
    Filter and evolve pseudo labels by Eq. \ref{eq:m_gt}\;
    Calculate semantic matching loss $\mathcal{L}_{SM}$ by Eq. \ref{eq:l_sm}\;
    {\color{teal}\# Distributional Matching} \\
    Calculate patch similarity by Eq. \ref{eq:gamma}\;
    Calculate distributional matching loss $\mathcal{L}_{DM}$ by Eq. \ref{eq:l_dm}\;
    {\color{teal}\# Image Updating} \\
    Combine two losses to obtain $\mathcal{L}_{IS}$ by Eq. \ref{eq:L_IS}\;
    Update $I_S$ by back-propagation of $\mathcal{L}_{IS}$\;
    }
    {\color{teal}\# \textbf{Stage 2: Quantization Calibration}}\\
    Input $I_S$ to $\mathcal{Q}$ and forward propagate $\mathcal{Q}(I_S)$\;
    Calibrate the initial quantization parameters by Eq. \ref{eq:sz}\;
    Perform scale reparameterization by Eq. \ref{eq:raparm}\;
\end{algorithm}

\subsection{The Overall Quantization Pipeline}
The quantization process of DFQ-SAM is divided into two stages: (i) image synthesis, and (ii) quantization calibration. The whole pipeline is summarized in Algorithm \ref{alg:1}. Each stage is described in detail below.

\noindent\textbf{Image Synthesis}
Our goal is to fully exploit the prior information in the pre-trained SAM, and thus facilitate the matching of synthesized images with real images from both semantic and distributional aspects. 
Note that $\mathcal{L}_{SM}$ and $\mathcal{L}_{DM}$ are specifically designed with careful consideration of the Transformer model architecture and the image segmentation task, thus resulting in significant performance.
Specifically, we take the image as the optimization objective and update it by calculating the following loss function:
\begin{equation}
    \mathcal{L}_{IS} = \mathcal{L}_{SM} + \beta \cdot \mathcal{L}_{DM},
    \label{eq:L_IS}
\end{equation}
where $\beta$ is the balance coefficient.

\noindent\textbf{Quantization Calibration}
After obtaining the synthesized images, we utilize them instead of the real dataset to calibrate the quantization parameters. In this stage, we employ scale reparameterization presented in Eq. \ref{eq:raparm} to ensure the accuracy of low-bit quantization. Note that we adopt post-training quantization, which does not require any resource-intensive training or fine-tuning, allowing the compression process to be achieved easily and quickly.

\section{Experiments}
We conduct exhaustive experiments to demonstrate the effectiveness and advantages of DFQ-SAM. The implementation details, including the models, datasets, comparison methods, and settings, are presented in Appendix \ref{app:exp_details}. In the following, we will discuss the experimental results in detail.

\begin{figure}[t]
\centering
\includegraphics[width=1.0\linewidth]{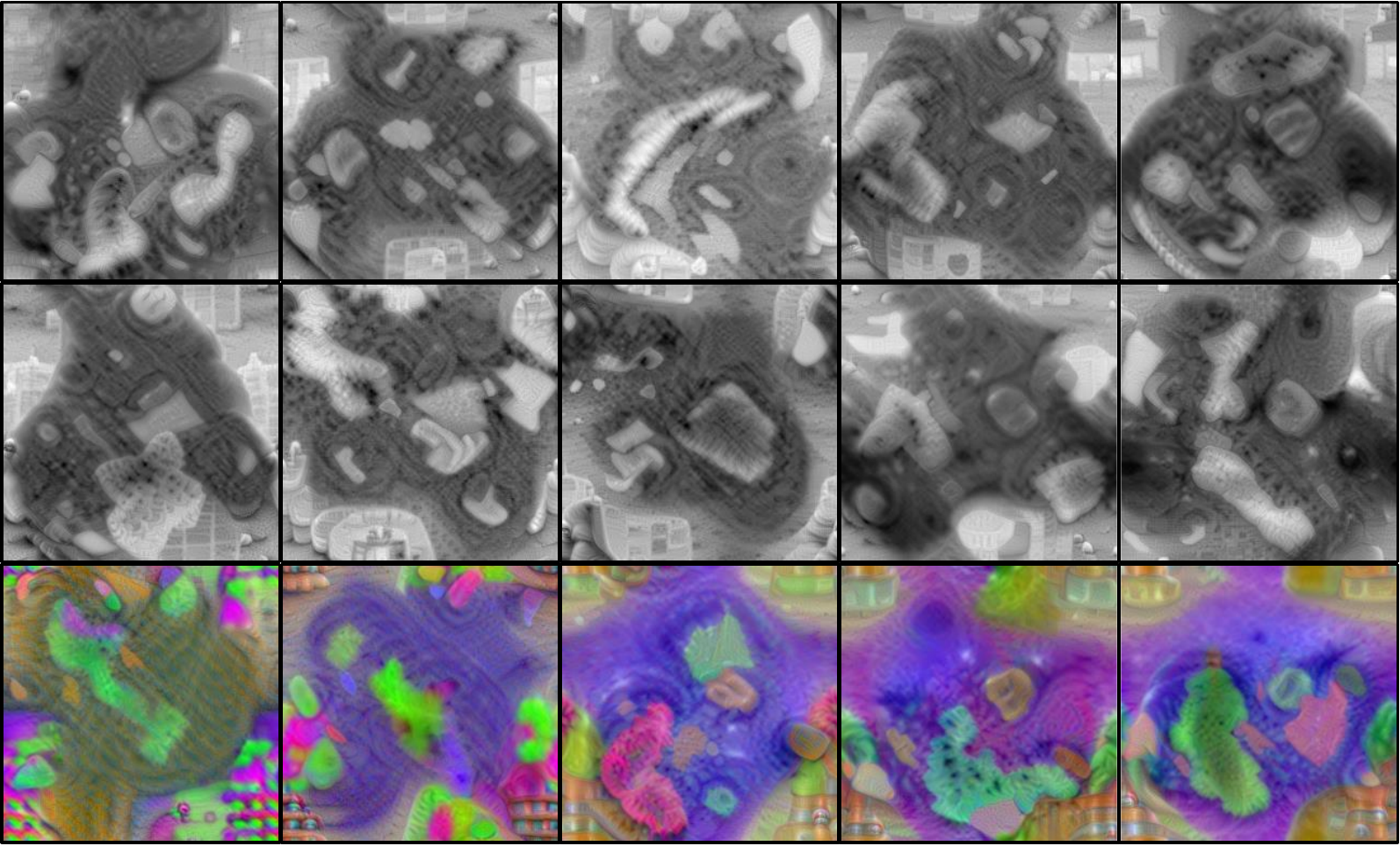}
\caption{Visualization of synthesized images (256$\times$256 pixels). The first two rows are grayscale images and the third row is color images, both reflecting medical semantic contents.}
\label{fig:image_vis}
\end{figure}

\begin{figure}[t]
\centering
\includegraphics[width=0.85\linewidth]{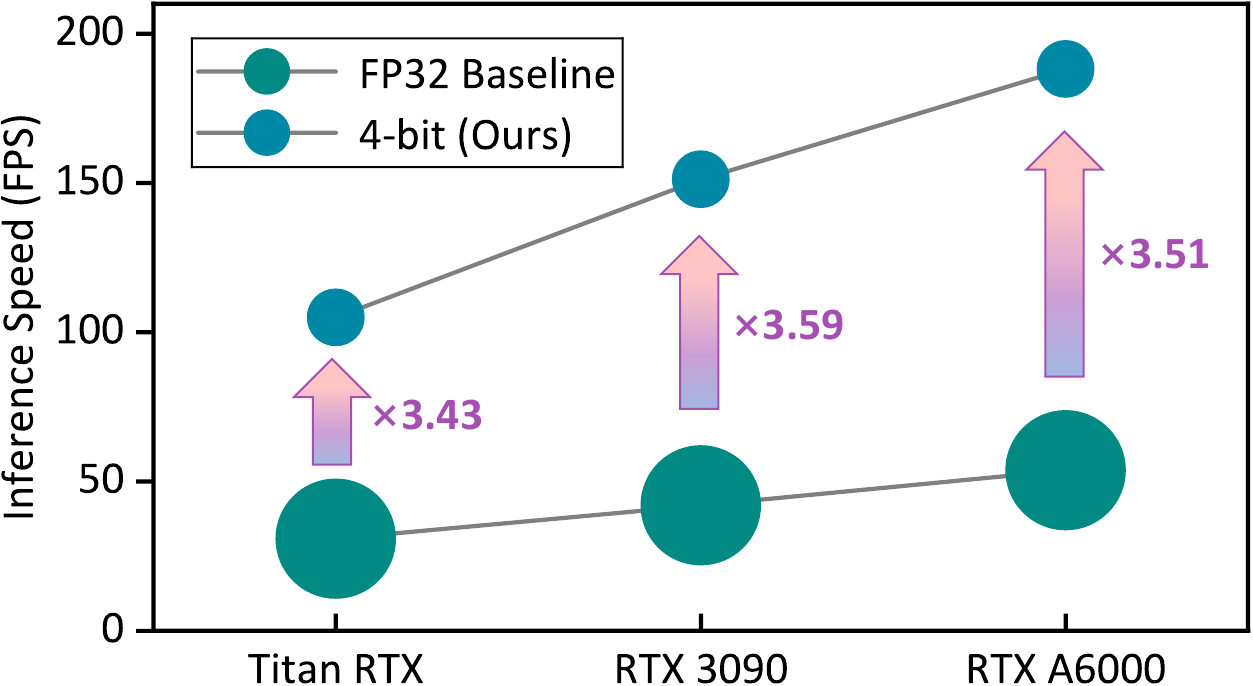}
\caption{Illustration of inference acceleration using 4-bit quantization across various devices, demonstrating a speedup ranging from 3.43$\times$ to 3.59$\times$ compared to the FP32 baseline.}
\label{fig:latency}
\end{figure}

\subsection{Analysis of Synthesized Images}
Fig. \ref{fig:image_vis} shows the visualization results of synthesized images (256$\times$256 pixels).
It should be emphasized that only the pre-trained model is required to synthesize these images, without any other information, especially the original data or any external prior.
In terms of brightness and contrast, we can observe that the synthesized images are well-aligned with the distribution of real images, demonstrating that the image synthesis process has successfully captured the fundamental visual priors in the pre-trained model. Additionally, from a semantic perspective, the synthesized images accurately reflect the anatomical structures and relative positions of various organs, aligning well with the characteristics of real images.
This alignment in both aspects plays a crucial role in downstream quantization calibration.

\subsection{Efficiency Evaluation of Quantized Models}
We demonstrate the effect of inference acceleration by deploying the 4-bit quantized model on different NVIDIA GPUs, including Titan RTX, RTX 3090, and RTX A6000.
In this implementation, the kernel of the quantized computation is built on CUTLASS \cite{kerr2017cutlass}, which is a high-performance CUDA library specifically designed for matrix multiplication. The acceleration results are presented in Fig. \ref{fig:latency}, indicating a 3.43$\sim$3.59$\times$ speedup compared to the full-precision baseline.
Therefore, the quantized SAM consistently delivers significantly enhanced inference efficiency across various hardware platforms, exhibiting strong compatibility and versatility. This ensures real-time and efficient medical services at the edge, which is particularly crucial in distributed systems with diverse hardware environments at local nodes, thereby promoting the widespread deployment and application of intelligent healthcare.

\subsection{Accuracy Evaluation of Quantized Models}
Here, we conduct extensive experiments on a variety set of datasets to demonstrate the accuracy advantages of the proposed DFQ-SAM. Notably, to highlight the generalization and robustness of the algorithm, the adopted datasets are from diverse modalities, including CT, MRI, PET, Ultrasound, Xray, Dermoscopy, and Endoscopy.
The evaluation metric is the predicted IoU of the masks. The accuracy results are presented and discussed in detail below.

\begin{table*}[t]\small
\setlength{\tabcolsep}{3pt}
\caption{Quantization results on grayscale medical images, with modalities of CT, MRI, PET, Ultrasound, and X-ray. DFQ-SAM is compared to methods that use real data or Gaussian noise for quantization. The reported values are the IoU results of segmentation masks on different datasets. We also present the model complexity, including model size and computational load (BOPs).}
\begin{tabular}{@{}ccccccccc@{}}
\toprule
\multirow{2.5}{*}{Method} &
\multirow{2.5}{*}{Privacy} &
  \multirow{2.5}{*}{Prec. (W/A)} &
  \multirow{2.5}{*}{Size (MB)} &
  \multirow{2.5}{*}{BOPs (G)} &
  \multicolumn{3}{c}{CT} &
  \multirow{2.5}{*}{Average} \\ \cmidrule(lr){6-8}
               &  &       &      &       & AbdomenCT1k\cite{Ma-2021-AbdomenCT-1K} & FLARE\cite{ma2023fast} & Synapse\cite{synapse} \\ \midrule
FP Baseline    & -   & 32/32 & 358  & 19866 &   76.93     & 75.58    & 76.34 & 76.28\\
Real Data      & \cm & 4/4   & 46.5 & 346   &   69.83     & 73.31    & 74.55 & 72.56\\
Gaussian Noise & \xm & 4/4   & 46.5 & 346   &   62.03     & 71.25    & 71.66 & 68.31 \\
\hc \begin{tabular}[c]{@{}c@{}} Synthesized Data\\ (\textbf{DFQ-SAM, Ours})\end{tabular} & \xm  &
  4/4 & 46.5 & 346 & 74.32  & 73.90  & 75.12 & 74.45 \\
  \midrule \midrule
\multirow{2.5}{*}{Method} &
\multirow{2.5}{*}{Privacy} &
  \multirow{2.5}{*}{Prec. (W/A)} &
  \multirow{2.5}{*}{Size (MB)} &
  \multirow{2.5}{*}{BOPs (G)} &
  \multicolumn{3}{c}{MRI} &
  \multirow{2.5}{*}{Average} \\ \cmidrule(lr){6-8}
               &  &       &      &       & ACDC\cite{bernard2018deep} & MSD\cite{antonelli2022medical} & ATLAS\cite{quinton2023tumour} \\ \midrule
FP Baseline    & -   & 32/32 & 358  & 19866 & 72.37     & 72.87    & 66.86 & 70.70\\
Real Data      & \cm & 4/4   & 46.5 & 346   & 69.92     & 71.62    & 63.42 & 68.32\\
Gaussian Noise & \xm & 4/4   & 46.5 & 346   & 67.54     & 70.44    & 59.69 & 65.89\\
\hc \begin{tabular}[c]{@{}c@{}} Synthesized Data\\ (\textbf{DFQ-SAM, Ours})\end{tabular} & \xm  &
  4/4 & 46.5 & 346 & 71.29 & 71.86 & 64.44 & 69.20\\
   \midrule \midrule
\multirow{2.5}{*}{Method} &
\multirow{2.5}{*}{Privacy} &
  \multirow{2.5}{*}{Prec. (W/A)} &
  \multirow{2.5}{*}{Size (MB)} &
  \multirow{2.5}{*}{BOPs (G)} &
  PET & Ultrasound & Xray &
  \multirow{2.5}{*}{Average} \\ \cmidrule(lr){6-6} \cmidrule(lr){7-7} \cmidrule(lr){8-8}
               &  &       &      &       & AutoPET\cite{gatidis2022whole} & SA-Ultrasound\cite{cheng2023sam} & SA-Xray\cite{cheng2023sam}  \\ \midrule
FP Baseline    & -   & 32/32 & 358  & 19866 & 64.10      & 69.53    & 47.63 & 60.42\\
Real Data      & \cm & 4/4   & 46.5 & 346   & 64.11      & 69.18    & 47.70 & 60.33\\
Gaussian Noise & \xm & 4/4   & 46.5 & 346   & 62.84      & 68.36    & 44.35 & 58.52\\
\hc \begin{tabular}[c]{@{}c@{}} Synthesized Data\\ (\textbf{DFQ-SAM, Ours})\end{tabular} & \xm  &
  4/4 & 46.5 & 346 & 65.46  & 68.99 & 45.47 & 59.97\\
   \bottomrule
\end{tabular}
\label{exp:gray_results}
\end{table*}

\begin{table*}[t]\small
\setlength{\tabcolsep}{4pt}
\caption{Quantization results on color medical images, with modalities of Dermoscopy and Endoscopy. DFQ-SAM is compared to methods that use real data or Gaussian noise for quantization. The reported values are the IoU results of segmentation masks on different datasets. We also present the model complexity, including model size and computational load (BOPs).}
\begin{tabular}{@{}ccccccccc@{}}
\toprule
\multirow{2.5}{*}{Method} &
\multirow{2.5}{*}{Privacy} &
  \multirow{2.5}{*}{Prec. (W/A)} &
  \multirow{2.5}{*}{Size (MB)} &
  \multirow{2.5}{*}{BOPs (G)} &
  Dermoscopy & \multicolumn{2}{c}{Endoscopy} &
  \multirow{2.5}{*}{Average} \\ \cmidrule(lr){6-6}  \cmidrule(lr){7-8}
               &  &       &      &       & ISIC\cite{gutman2016skin}\cite{codella2018skin} & EndoVis\cite{allan20192017} & SA-Endoscopy\cite{cheng2023sam} \\ \midrule
FP Baseline    & -   & 32/32 & 358  & 19866 & 84.24     & 70.44    & 83.38 & 79.35\\
Real Data      & \cm & 4/4   & 46.5 & 346   & 82.71     & 66.31    & 78.91 & 75.98\\
Gaussian Noise & \xm & 4/4   & 46.5 & 346   & 80.59     & 63.82    & 76.09 & 73.50\\
\hc \begin{tabular}[c]{@{}c@{}} Synthesized Data\\ (\textbf{DFQ-SAM, Ours})\end{tabular} & \xm  &
  4/4 & 46.5 & 346 & 83.23  & 66.23 & 79.01 & 76.16\\
   \bottomrule
\end{tabular}
\label{exp:color_results}
\end{table*}

\noindent\textbf{Main Results}
Tables \ref{exp:gray_results} and \ref{exp:color_results} report the quantization results for image segmentation on grayscale and color datasets, respectively.
They also report the model complexity, including model size and computational load (BOPs), where the former is reduced by $\sim$8$\times$ and the latter by $\sim$64$\times$. 
We begin by discussing the results on grayscale datasets. Calibration using Gaussian noise apparently results in unsatisfactory performance; in contrast, DFQ-SAM achieves significantly improved performance by utilizing synthesized data.
For instance, in the CT modality, the average accuracy of DFQ-SAM in 4-bit quantization on the three datasets AbdomenCT1k, FLARE, and Synapse is 74.45, which is only a 1.83 degradation from the full-precision baseline. 
DFQ-SAM also achieves encouraging performance on the datasets of MRI modalities ACDC, MSD, and ATLAS, with an average accuracy of 69.20, producing only 1.50 accuracy loss.
Moreover, the performance of DFQ-SAM remains robust on color datasets. For ISIC and SA-Endoscopy datasets, the accuracy of DFQ-SAM reaches 83.23 and 79.01, respectively, which is even higher than the real-data method, underscoring DFQ-SAM's capability to maintain high accuracy.

\begin{figure*}[t]
\centering
\subfigure[DFQ-SAM vs. Real-data Quant. ]{
\includegraphics[width=0.55\linewidth]{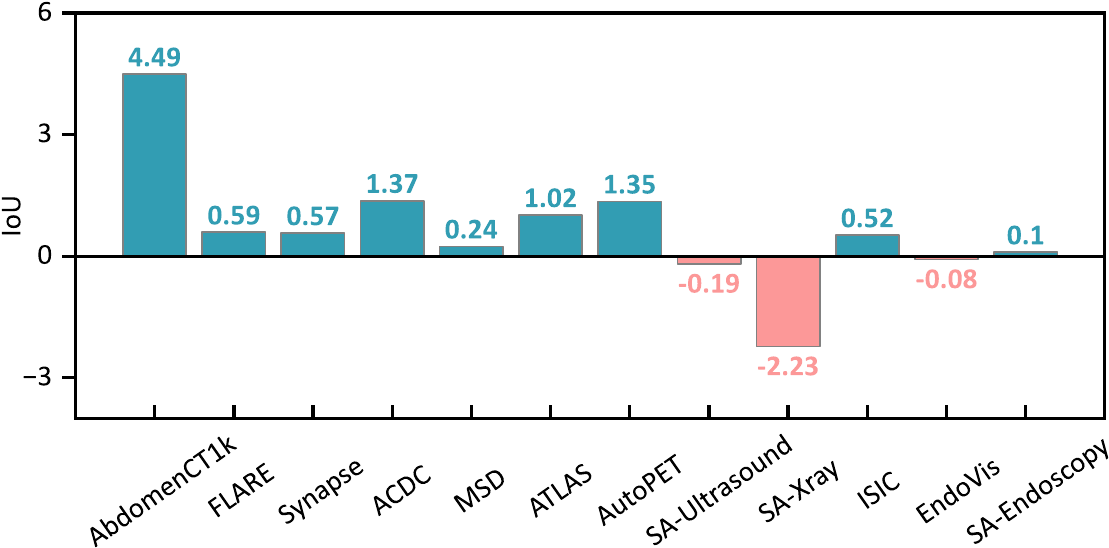}
\label{fig:exp_bar}
} \qquad
\subfigure[DFQ-SAM vs. FP Baseline]{
\includegraphics[width=0.35\linewidth]{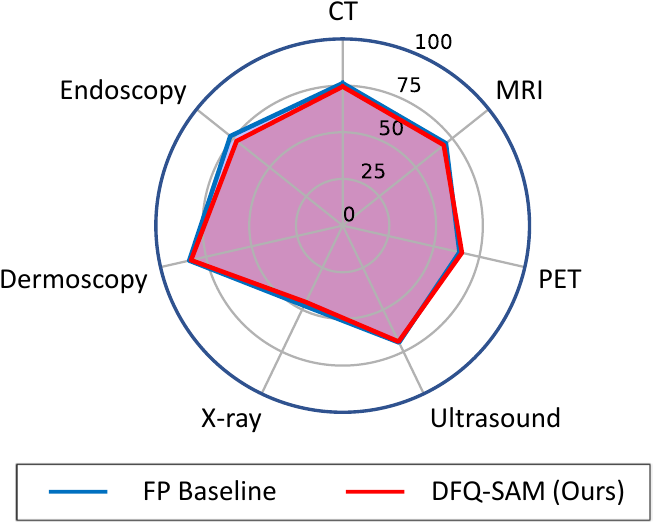}
\label{fig:exp_radar}
} 
\caption{Illustration of comparisons of quantization performance. (a) is a bar chart of the 4-bit performance differences between DFQ-SAM and real-data quantization on various datasets, which highlights that DFQ-SAM using synthesized data can outperform the method using real data in most cases. (b) is a performance radargram of DFQ-SAM versus the full-precision baseline across various image modalities, demonstrating that 4-bit DFQ-SAM can achieve comparable performance.}
\label{fig:exp_bar_radar}
\end{figure*}

\noindent\textbf{DFQ-SAM vs. Real-data Quant.}
We also present a more intuitive comparison between DFA-SAM and the real-data method, as shown in Fig. \ref{fig:exp_bar}. The values are the difference in performance between the two, and a value greater than 0 is a case where DFQ-SAM outperforms the real-data method. Surprisingly, in most cases, DFQ-SAM is on the superior side, which means that the synthesized data is more effective for calibration than the real data. This is because the synthesized images learn the prior from the whole dataset and represent the comprehensive distribution of the overall dataset, and therefore have a better information capacity than the real data that represents individual instances.

\noindent\textbf{DFQ-SAM vs. FP Baseline}
Fig. \ref{fig:exp_radar} illustrates the performance radargram of DFQ-SAM and the full-precision baseline. Intuitively, DFQ-SAM, despite adopting low-bit quantization, continues to deliver performance that closely matches that of the high-precision pre-trained model across various datasets, effectively preserving critical features even with reduced bit precision.
Therefore, DFQ-SAM is proven capable of achieving near-lossless compression, fully demonstrating its reliability in real-world applications.

\begin{figure}[t]
\centering
\subfigure[AbdomenCT1k Dataset]{
\includegraphics[width=0.8\linewidth]{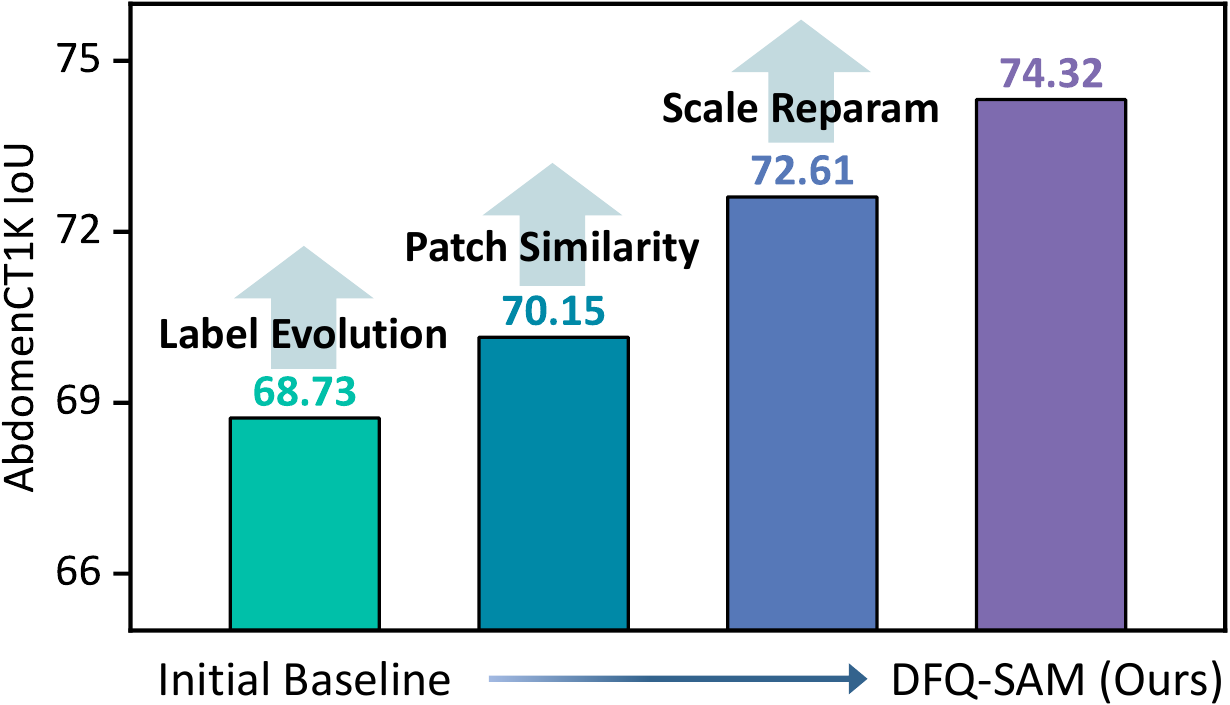}
} \\
\subfigure[FLARE Dataset]{
\includegraphics[width=0.8\linewidth]{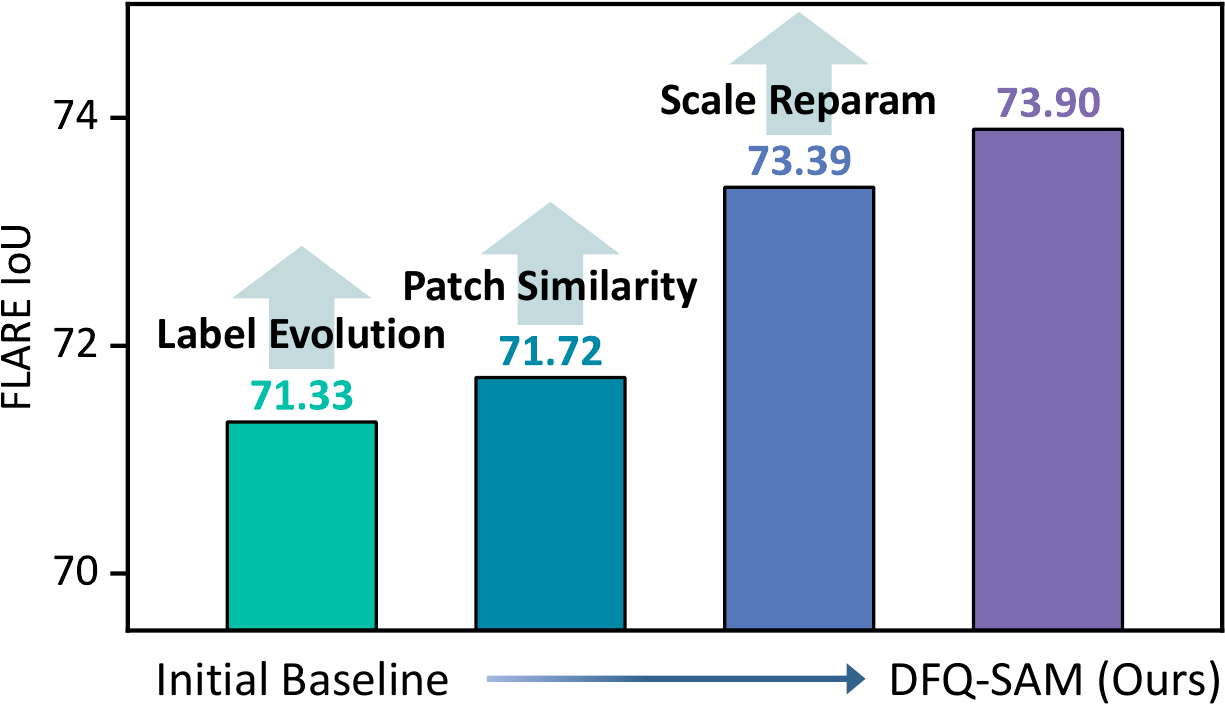}
}
\caption{Ablation experiments on the effectiveness of each component in DFQ-SAM. Each component contributes significantly, resulting in impressive performance of DFQ-SAM.}
\label{fig:exp_ablation}
\end{figure}

\begin{figure}[t]
\centering
\subfigure[Grid Search for $\alpha$]{
\includegraphics[width=0.89\linewidth]{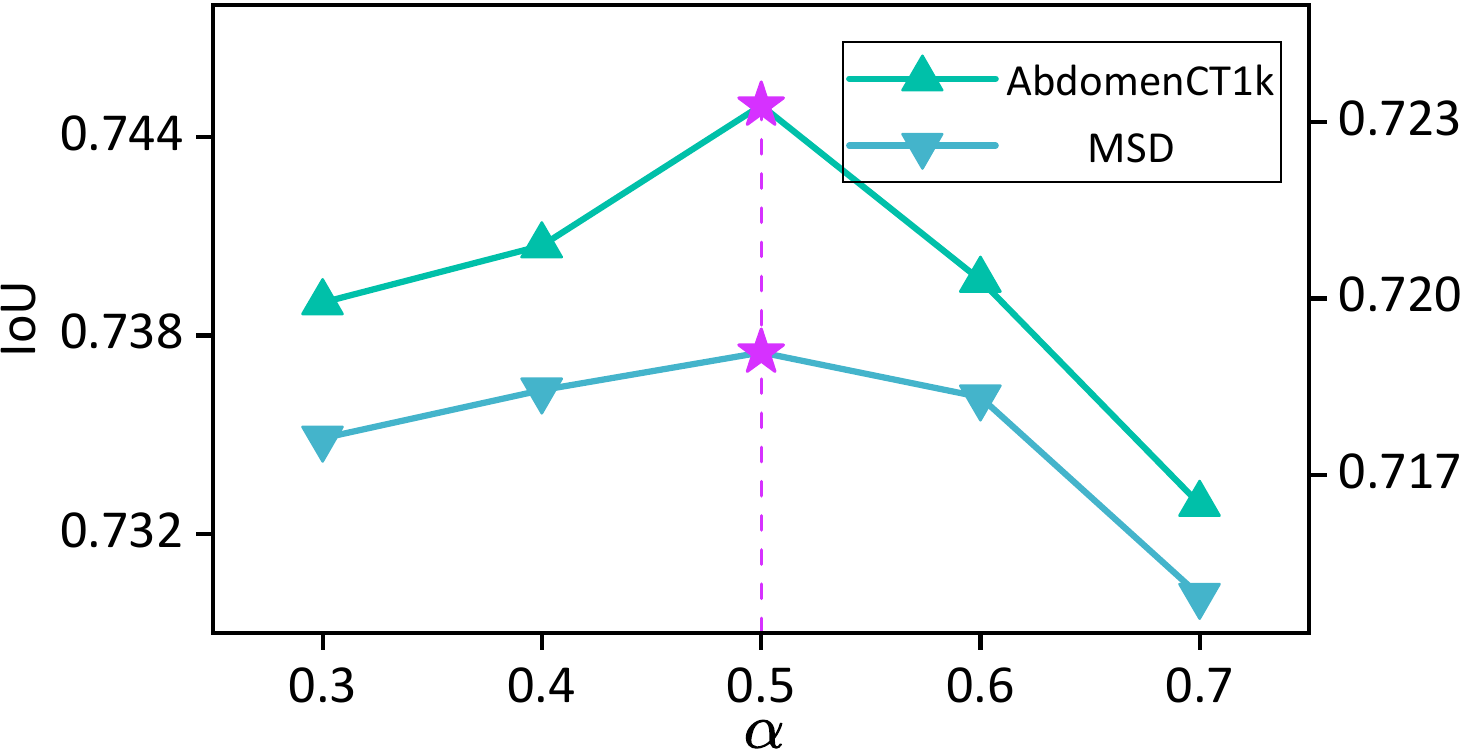}
} \\
\subfigure[Grid Search for $\beta$]{
\includegraphics[width=0.89\linewidth]{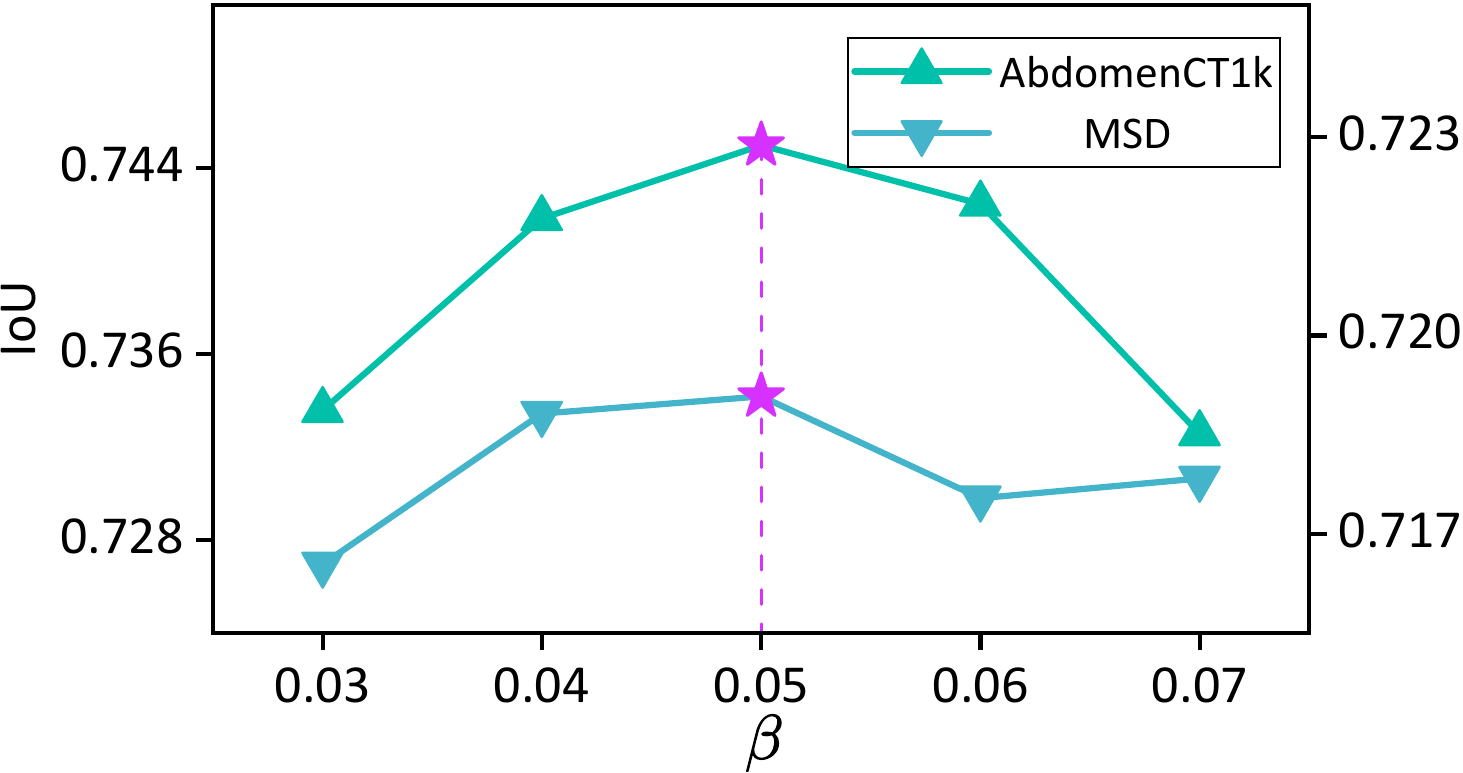}
}
\caption{Grid search for balance hyperparameters $\alpha$ and $\beta$. The quantization performance is robust to hyperparameters, and their optimal values remain consistent across different datasets.}
\label{fig:exp_grid_search}
\end{figure}

\subsection{Ablation Studies}
\noindent\textbf{Effectiveness of Each Component}
Here, we verify the effectiveness of each component in DFQ-SAM, and the results are shown in Fig. \ref{fig:exp_ablation}. The initial baseline is established by optimizing Gaussian noise with random labels.
Building on this, the sequential addition of the proposed modules, including label evolution, patch similarity, and scale reparameterization, further enhances performance at each step. For instance, on AbdomenCT1k dataset, the initial accuracy is 68.73, and the improvements by the three components are 1.42, 2.46, and 1.71, respectively, resulting in a final accuracy of 74.32. Therefore, each component plays a crucial role in improving the overall performance.

\noindent\textbf{Grid Search for Hyperparameters}
We also assess the impact of the balance hyperparameters $\alpha$ and $\beta$ through a grid search and determine the optimal values. The results are shown in Fig. \ref{fig:exp_grid_search}. 
We observe that the quantization performance is robust to changes in hyperparameters, showing only minor variations, such as between 0.732 and 0.744. More importantly, they reach the optimal value at the same point (0.5, 0.05) across different datasets.

\section{Conclusion}
In this paper, we propose DFQ-SAM, a model quantization framework for medical SAM without requiring any original training data, thus preserving data privacy in model compression and deployment.
DFQ-SAM effectively leverages the prior knowledge from the pre-trained model to invert the synthesized data, aligning it with real data in both semantic and distributional aspects. It then employs an advanced quantization calibration strategy, achieving near lossless 4-bit model compression. To the best of our knowledge, this is the first successful attempt at data-free quantization for SAM. Exhaustive experiments demonstrate that DFQ-SAM exhibits encouraging performance on a range of datasets with multiple modalities, even surpassing the real-data-driven approach.
With the proposed framework, the model complexity of medical SAM is significantly reduced, while data privacy is well protected in cloud-side collaboration, which facilitates the provision of safe, reliable, and fast intelligent medical services on resource-constrained devices. This helps to actively promote the downward sinking of high-quality medical resources, ensuring more equitable access to healthcare in remote and underdeveloped areas, which is crucial for addressing the current imbalance in the distribution of medical resources.


{
\bibliography{egbib}

\begin{thebibliography}{10}

\bibitem{pramesh2022priorities}
C.~Pramesh, R.~A. Badwe, N.~Bhoo-Pathy, C.~M. Booth, G.~Chinnaswamy, A.~J. Dare, V.~P. de~Andrade, D.~J. Hunter, S.~Gopal, M.~Gospodarowicz, {\em et~al.}, ``Priorities for cancer research in low-and middle-income countries: a global perspective,'' {\em Nature medicine}, vol.~28, no.~4, pp.~649--657, 2022.

\bibitem{brito2022global}
A.~F. Brito, E.~Semenova, G.~Dudas, G.~W. Hassler, C.~C. Kalinich, M.~U. Kraemer, J.~Ho, H.~Tegally, G.~Githinji, C.~N. Agoti, {\em et~al.}, ``Global disparities in sars-cov-2 genomic surveillance,'' {\em Nature communications}, vol.~13, no.~1, p.~7003, 2022.

\bibitem{gelband2016costs}
H.~Gelband, R.~Sankaranarayanan, C.~L. Gauvreau, S.~Horton, B.~O. Anderson, F.~Bray, J.~Cleary, A.~J. Dare, L.~Denny, M.~K. Gospodarowicz, {\em et~al.}, ``Costs, affordability, and feasibility of an essential package of cancer control interventions in low-income and middle-income countries: key messages from disease control priorities,'' {\em The Lancet}, vol.~387, no.~10033, pp.~2133--2144, 2016.

\bibitem{chalkidou2014evidence}
K.~Chalkidou, P.~Marquez, P.~K. Dhillon, Y.~Teerawattananon, T.~Anothaisintawee, C.~A.~G. Gadelha, and R.~Sullivan, ``Evidence-informed frameworks for cost-effective cancer care and prevention in low, middle, and high-income countries,'' {\em The lancet oncology}, vol.~15, no.~3, pp.~e119--e131, 2014.

\bibitem{ma2024segment}
J.~Ma, Y.~He, F.~Li, L.~Han, C.~You, and B.~Wang, ``Segment anything in medical images,'' {\em Nature Communications}, vol.~15, no.~1, p.~654, 2024.

\bibitem{cheng2023sam}
J.~Cheng, J.~Ye, Z.~Deng, J.~Chen, T.~Li, H.~Wang, Y.~Su, Z.~Huang, J.~Chen, L.~Jiang, {\em et~al.}, ``Sam-med2d,'' {\em arXiv preprint arXiv:2308.16184}, 2023.

\bibitem{wei2023medsam}
X.~Wei, J.~Cao, Y.~Jin, M.~Lu, G.~Wang, and S.~Zhang, ``I-medsam: Implicit medical image segmentation with segment anything,'' {\em arXiv preprint arXiv:2311.17081}, 2023.

\bibitem{lv2024ptq4sam}
C.~Lv, H.~Chen, J.~Guo, Y.~Ding, and X.~Liu, ``Ptq4sam: Post-training quantization for segment anything,'' in {\em Proceedings of the IEEE/CVF Conference on Computer Vision and Pattern Recognition}, pp.~15941--15951, 2024.

\bibitem{zhang2023faster}
C.~Zhang, D.~Han, Y.~Qiao, J.~U. Kim, S.-H. Bae, S.~Lee, and C.~S. Hong, ``Faster segment anything: Towards lightweight sam for mobile applications,'' {\em arXiv preprint arXiv:2306.14289}, 2023.

\bibitem{gholami2022survey}
A.~Gholami, S.~Kim, Z.~Dong, Z.~Yao, M.~W. Mahoney, and K.~Keutzer, ``A survey of quantization methods for efficient neural network inference,'' in {\em Low-Power Computer Vision}, pp.~291--326, Chapman and Hall/CRC, 2022.

\bibitem{krishnamoorthi2018quantizing}
R.~Krishnamoorthi, ``Quantizing deep convolutional networks for efficient inference: A whitepaper,'' {\em arXiv preprint arXiv:1806.08342}, 2018.

\bibitem{wang2019haq}
K.~Wang, Z.~Liu, Y.~Lin, J.~Lin, and S.~Han, ``Haq: Hardware-aware automated quantization with mixed precision,'' in {\em Proceedings of the IEEE/CVF conference on computer vision and pattern recognition}, pp.~8612--8620, 2019.

\bibitem{esser2019learned}
S.~K. Esser, J.~L. McKinstry, D.~Bablani, R.~Appuswamy, and D.~S. Modha, ``Learned step size quantization,'' {\em arXiv preprint arXiv:1902.08153}, 2019.

\bibitem{li2024htq}
Z.~Li, X.~Long, J.~Xiao, and Q.~Gu, ``Htq: Exploring the high-dimensional trade-off of mixed-precision quantization,'' {\em Pattern Recognition}, p.~110788, 2024.

\bibitem{cai2020zeroq}
Y.~Cai, Z.~Yao, Z.~Dong, A.~Gholami, M.~W. Mahoney, and K.~Keutzer, ``Zeroq: A novel zero shot quantization framework,'' in {\em Proceedings of the IEEE/CVF conference on computer vision and pattern recognition}, pp.~13169--13178, 2020.

\bibitem{zhong2022intraq}
Y.~Zhong, M.~Lin, G.~Nan, J.~Liu, B.~Zhang, Y.~Tian, and R.~Ji, ``Intraq: Learning synthetic images with intra-class heterogeneity for zero-shot network quantization,'' in {\em Proceedings of the IEEE/CVF Conference on Computer Vision and Pattern Recognition}, pp.~12339--12348, 2022.

\bibitem{xu2020generative}
S.~Xu, H.~Li, B.~Zhuang, J.~Liu, J.~Cao, C.~Liang, and M.~Tan, ``Generative low-bitwidth data free quantization,'' in {\em Computer Vision--ECCV 2020: 16th European Conference, Glasgow, UK, August 23--28, 2020, Proceedings, Part XII 16}, pp.~1--17, Springer, 2020.

\bibitem{zhang2021diversifying}
X.~Zhang, H.~Qin, Y.~Ding, R.~Gong, Q.~Yan, R.~Tao, Y.~Li, F.~Yu, and X.~Liu, ``Diversifying sample generation for accurate data-free quantization,'' in {\em Proceedings of the IEEE/CVF conference on computer vision and pattern recognition}, pp.~15658--15667, 2021.

\bibitem{li2022dual}
Z.~Li, L.~Ma, X.~Long, J.~Xiao, and Q.~Gu, ``Dual-discriminator adversarial framework for data-free quantization,'' {\em Neurocomputing}, vol.~511, pp.~67--77, 2022.

\bibitem{fan2024data}
C.~Fan, Z.~Wang, D.~Guo, and M.~Wang, ``Data-free quantization via pseudo-label filtering,'' in {\em Proceedings of the IEEE/CVF Conference on Computer Vision and Pattern Recognition}, pp.~5589--5598, 2024.

\bibitem{dosovitskiy2020image}
A.~Dosovitskiy, L.~Beyer, A.~Kolesnikov, D.~Weissenborn, X.~Zhai, T.~Unterthiner, M.~Dehghani, M.~Minderer, G.~Heigold, S.~Gelly, {\em et~al.}, ``An image is worth 16x16 words: Transformers for image recognition at scale,'' {\em arXiv preprint arXiv:2010.11929}, 2020.

\bibitem{li2022patch}
Z.~Li, L.~Ma, M.~Chen, J.~Xiao, and Q.~Gu, ``Patch similarity aware data-free quantization for vision transformers,'' in {\em European conference on computer vision}, pp.~154--170, Springer, 2022.

\bibitem{li2023psaq}
Z.~Li, M.~Chen, J.~Xiao, and Q.~Gu, ``Psaq-vit v2: Toward accurate and general data-free quantization for vision transformers,'' {\em IEEE Transactions on Neural Networks and Learning Systems}, 2023.

\bibitem{li2023repq}
Z.~Li, J.~Xiao, L.~Yang, and Q.~Gu, ``Repq-vit: Scale reparameterization for post-training quantization of vision transformers,'' in {\em Proceedings of the IEEE/CVF International Conference on Computer Vision}, pp.~17227--17236, 2023.

\bibitem{patil2013medical}
D.~D. Patil and S.~G. Deore, ``Medical image segmentation: a review,'' {\em International Journal of Computer Science and Mobile Computing}, vol.~2, no.~1, pp.~22--27, 2013.

\bibitem{wang2021annotation}
S.~Wang, C.~Li, R.~Wang, Z.~Liu, M.~Wang, H.~Tan, Y.~Wu, X.~Liu, H.~Sun, R.~Yang, {\em et~al.}, ``Annotation-efficient deep learning for automatic medical image segmentation,'' {\em Nature communications}, vol.~12, no.~1, p.~5915, 2021.

\bibitem{zhou2018unet++}
Z.~Zhou, M.~M. Rahman~Siddiquee, N.~Tajbakhsh, and J.~Liang, ``Unet++: A nested u-net architecture for medical image segmentation,'' in {\em Deep Learning in Medical Image Analysis and Multimodal Learning for Clinical Decision Support: 4th International Workshop, DLMIA 2018, and 8th International Workshop, ML-CDS 2018, Held in Conjunction with MICCAI 2018, Granada, Spain, September 20, 2018, Proceedings 4}, pp.~3--11, Springer, 2018.

\bibitem{chen2021transunet}
J.~Chen, Y.~Lu, Q.~Yu, X.~Luo, E.~Adeli, Y.~Wang, L.~Lu, A.~L. Yuille, and Y.~Zhou, ``Transunet: Transformers make strong encoders for medical image segmentation,'' {\em arXiv preprint arXiv:2102.04306}, 2021.

\bibitem{kirillov2023segment}
A.~Kirillov, E.~Mintun, N.~Ravi, H.~Mao, C.~Rolland, L.~Gustafson, T.~Xiao, S.~Whitehead, A.~C. Berg, W.-Y. Lo, {\em et~al.}, ``Segment anything,'' in {\em Proceedings of the IEEE/CVF International Conference on Computer Vision}, pp.~4015--4026, 2023.

\bibitem{Wang2023SAMMed3D}
H.~Wang, S.~Guo, J.~Ye, Z.~Deng, J.~Cheng, T.-X. Li, J.~Chen, Y.-C. Su, Z.~Huang, Y.~Shen, B.~Fu, S.~Zhang, J.~He, and Y.~Qiao, ``Sam-med3d,'' {\em arXiv preprint arXiv:2310.15161}, 2023.

\bibitem{zhao2023fast}
X.~Zhao, W.~Ding, Y.~An, Y.~Du, T.~Yu, M.~Li, M.~Tang, and J.~Wang, ``Fast segment anything,'' {\em arXiv preprint arXiv:2306.12156}, 2023.

\bibitem{li2023vit}
Z.~Li and Q.~Gu, ``I-vit: Integer-only quantization for efficient vision transformer inference,'' in {\em Proceedings of the IEEE/CVF International Conference on Computer Vision}, pp.~17065--17075, 2023.

\bibitem{li2024repquant}
Z.~Li, X.~Liu, J.~Zhang, and Q.~Gu, ``Repquant: Towards accurate post-training quantization of large transformer models via scale reparameterization,'' {\em arXiv preprint arXiv:2402.05628}, 2024.

\bibitem{chawla2021data}
A.~Chawla, H.~Yin, P.~Molchanov, and J.~Alvarez, ``Data-free knowledge distillation for object detection,'' in {\em Proceedings of the IEEE/CVF Winter Conference on Applications of Computer Vision}, pp.~3289--3298, 2021.

\bibitem{kerr2017cutlass}
A.~Kerr, D.~Merrill, J.~Demouth, and J.~Tran, ``Cutlass: Fast linear algebra in cuda c++,'' {\em NVIDIA Developer Blog}, 2017.

\bibitem{Ma-2021-AbdomenCT-1K}
J.~Ma, Y.~Zhang, S.~Gu, C.~Zhu, C.~Ge, Y.~Zhang, X.~An, C.~Wang, Q.~Wang, X.~Liu, S.~Cao, Q.~Zhang, S.~Liu, Y.~Wang, Y.~Li, J.~He, and X.~Yang, ``Abdomenct-1k: Is abdominal organ segmentation a solved problem?,'' {\em IEEE Transactions on Pattern Analysis and Machine Intelligence}, vol.~44, no.~10, pp.~6695--6714, 2022.

\bibitem{ma2023fast}
J.~Ma and B.~Wang, {\em Fast and Low-Resource Semi-supervised Abdominal Organ Segmentation: MICCAI 2022 Challenge, FLARE 2022, Held in Conjunction with MICCAI 2022, Singapore, September 22, 2022, Proceedings}, vol.~13816.
\newblock Springer Nature, 2023.

\bibitem{synapse}
``Multi-atlas labeling beyond the cranial vault - workshop and challenge,'' 2015.

\bibitem{bernard2018deep}
O.~Bernard, A.~Lalande, C.~Zotti, F.~Cervenansky, X.~Yang, P.-A. Heng, I.~Cetin, K.~Lekadir, O.~Camara, M.~A.~G. Ballester, {\em et~al.}, ``Deep learning techniques for automatic mri cardiac multi-structures segmentation and diagnosis: is the problem solved?,'' {\em IEEE transactions on medical imaging}, vol.~37, no.~11, pp.~2514--2525, 2018.

\bibitem{antonelli2022medical}
M.~Antonelli, A.~Reinke, S.~Bakas, K.~Farahani, A.~Kopp-Schneider, B.~A. Landman, G.~Litjens, B.~Menze, O.~Ronneberger, R.~M. Summers, {\em et~al.}, ``The medical segmentation decathlon,'' {\em Nature communications}, vol.~13, no.~1, p.~4128, 2022.

\bibitem{quinton2023tumour}
F.~Quinton, R.~Popoff, B.~Presles, S.~Leclerc, F.~Meriaudeau, G.~Nodari, O.~Lopez, J.~Pellegrinelli, O.~Chevallier, D.~Ginhac, {\em et~al.}, ``A tumour and liver automatic segmentation (atlas) dataset on contrast-enhanced magnetic resonance imaging for hepatocellular carcinoma,'' {\em Data}, vol.~8, no.~5, p.~79, 2023.

\bibitem{gatidis2022whole}
S.~Gatidis, T.~Hepp, M.~Fr{\"u}h, C.~La~Foug{\`e}re, K.~Nikolaou, C.~Pfannenberg, B.~Sch{\"o}lkopf, T.~K{\"u}stner, C.~Cyran, and D.~Rubin, ``A whole-body fdg-pet/ct dataset with manually annotated tumor lesions,'' {\em Scientific Data}, vol.~9, no.~1, p.~601, 2022.

\bibitem{gutman2016skin}
D.~Gutman, N.~C. Codella, E.~Celebi, B.~Helba, M.~Marchetti, N.~Mishra, and A.~Halpern, ``Skin lesion analysis toward melanoma detection: A challenge at the international symposium on biomedical imaging (isbi) 2016, hosted by the international skin imaging collaboration (isic),'' {\em arXiv preprint arXiv:1605.01397}, 2016.

\bibitem{codella2018skin}
N.~C. Codella, D.~Gutman, M.~E. Celebi, B.~Helba, M.~A. Marchetti, S.~W. Dusza, A.~Kalloo, K.~Liopyris, N.~Mishra, H.~Kittler, {\em et~al.}, ``Skin lesion analysis toward melanoma detection: A challenge at the 2017 international symposium on biomedical imaging (isbi), hosted by the international skin imaging collaboration (isic),'' in {\em 2018 IEEE 15th international symposium on biomedical imaging (ISBI 2018)}, pp.~168--172, IEEE, 2018.

\bibitem{allan20192017}
M.~Allan, A.~Shvets, T.~Kurmann, Z.~Zhang, R.~Duggal, Y.-H. Su, N.~Rieke, I.~Laina, N.~Kalavakonda, S.~Bodenstedt, {\em et~al.}, ``2017 robotic instrument segmentation challenge,'' {\em arXiv preprint arXiv:1902.06426}, 2019.

\bibitem{kingma2014adam}
D.~P. Kingma and J.~Ba, ``Adam: A method for stochastic optimization,'' {\em arXiv preprint arXiv:1412.6980}, 2014.

\end{thebibliography}
\bibliographystyle{ieeetr}
}

\clearpage

\appendices
\section{Our preliminary efforts for DFQ-SAM}
Our preliminary efforts, PSAQ-ViT\cite{li2022patch,li2023psaq} and RepQ-ViT\cite{li2023repq}, have laid a strong foundation for the success of DFQ-SAM, allowing us to accomplish accurate low-bit model quantization without any original data. Here, we present their ideas and details.

\subsection{PSAQ-ViT: Patch Similarity Metric}
\label{app:pasq}

PSAQ-ViT is the first successful attempt within the community to quantize Transformers without any original data, aiming to utilize the inherent prior information embedded in the unique properties of Transformer models to synthesize data. In the absence of an absolute value metric such as BatchNorm statistics, the approach focus on assessing the general difference in model inference when the input is Gaussian noise versus real images, and anticipates developing a \emph{relative value} metric accordingly to optimize the noise by reducing the above difference.

As noted by \cite{dosovitskiy2020image}, the self-attention mechanism in Transformers is designed to extract key information during training, such as distinguishing the foreground from the background, which helps the model make accurate decisions. During the inference phase of a pre-trained model, real images generate varying responses between foreground and background patches, resulting in diverse patch similarities (i.e., differences in response similarity across patches). In contrast, Gaussian noise inputs, which do not clearly separate foreground from background, lead to more uniform responses.

Therefore, PSAQ-ViT treats the diversity of patch similarities as the key difference and quantifies this diversity using the differential entropy of patch responses in self-attention mechanism. By using this measure as an optimization criterion, the approach gradually transforms the image from Gaussian noise to one with increasingly diverse responses, thereby closely approximating a real image.

The patch similarity, its differential entropy, and the loss are presented in Eqs. \ref{eq:gamma}, \ref{eq:entropy}, and \ref{eq:l_dm}, respectively.

\subsection{RepQ-ViT: Scale Reparameterization}
\label{app:repq}
RepQ-ViT enables accurate low-bit (e.g., 4-bit) quantization calibration without the need for cumbersome and resource-intensive training or fine-tuning. It achieves this by elegantly addressing the challenge of quantizing components with extreme distributions. Specifically, it proposes a novel quantization-inference decoupling paradigm, where the former employs complex quantizers to maintain the data distribution as much as possible, and the latter employs simplified quantizers to perform fast inference. And importantly, the two can be equivalently transformed by scale reparameterization.

The proposed paradigm is remarkably effective for components with extreme distributions, especially LayerNorm activations with severe inter-channel variance. RepQ-ViT first performs channel-wise quantization for LayerNorm activations in the quantization process, i.e., each channel has a unique quantization scale, and then equivalently transforms to layer-wise quantization in the inference process, i.e., all channels share a unified quantization scale. In the following we describe the derivation of the equivalent transformations.

Given LayerNorm activations $\bm{X}$, when performing channel-wise quantization, the quantization scale $\bm{s}\in R^{D}$ and zero-point $\bm{z}\in Z^{D}$ are obtained.
The objective is to reparameterize them into scalars $\tilde{s}\in R^{1}$ and $\tilde{z}\in Z^{1}$ to accommodate layer-wise quantization.
Here, $\tilde{s}$ and $\tilde{z}$ are pre-specified and set to their respective mean values, i.e., $\tilde{s}=\text{E}[\bm{s}], \tilde{z}=\text{E}[\bm{z}]$.  Introducing the variation factors $\bm{r}_1=\bm{s}/\tilde{\bm{s}}$ and $\bm{r}_2=\bm{z}-\tilde{\bm{z}}$, the following equations are established:
\begin{align}
  \label{eq:3.2-1} \tilde{\bm{z}} &= \bm{z}-\bm{r}_2 = \left\lfloor -\frac{\left[\min(\bm{X}_{:,d})\right]_{1\leq d \leq D}+\bm{s}\odot \bm{r}_2}{\bm{s}} \right\rceil \\
  \label{eq:3.2-2} \tilde{\bm{s}} &= \frac{\bm{s}}{\bm{r}_1} = \frac{\left[\max(\bm{X}_{:,d})-\min(\bm{X}_{:,d})\right]_{1\leq d \leq D}/\bm{r}_1}{2^b-1}
\end{align}

Eq. \ref{eq:3.2-1} indicates that adding $\bm{s}\odot \bm{r}_2$ to each channel of $\bm{X}$ results in $\tilde{\bm{z}}$, while Eq. \ref{eq:3.2-2} shows that dividing each channel of $\bm{X}$ by $\bm{r}_1$ achieves $\tilde{\bm{s}}$. These adjustments can be implemented by modifying the affine factors in LayerNorm layer as follows:
\begin{equation}
    \label{eq:raparm_1}
  \widetilde{\bm{\beta}} = \frac{\bm{\beta}+\bm{s}\odot \bm{r}_2}{\bm{r}_1}, \quad \widetilde{\bm{\gamma}} = \frac{\bm{\gamma}}{\bm{r}_1}
\end{equation}

The above procedure results in a distribution shift of activations, i.e., $\widetilde{\bm{X}}_{n,:}=(\bm{X}_{n,:}+\bm{s}\odot \bm{r}_2)/\bm{r}_1$. This distribution shift can be corrected by applying inverse compensation to the weights of the subsequent layer. Specifically, using equivalent transformations, we obtain:
\begin{equation}
\begin{split}
  \bm{X}_{n,:}\bm{W}^{qkv}_{:,j}+\bm{b}^{qkv}_j = \frac{\bm{X}_{n,:}+\bm{s}\odot \bm{r}_2}{\bm{r}_1} \left(\bm{r}_1\odot\bm{W}^{qkv}_{:,j}\right) \\ + \left(\bm{b}^{qkv}_j - (\bm{s}\odot \bm{r}_2) \bm{W}^{qkv}_{:,j}\right)
\end{split}
\end{equation}
which suggests that to ensure the next layer's outputs remain consistent, we need to adjust the weights as follows:
\begin{equation}
\label{eq:raparm_2}
\begin{aligned}
  \widetilde{\bm{W}}^{qkv}_{:,j} &= \bm{r}_1\odot\bm{W}^{qkv}_{:,j} \\
  \widetilde{\bm{b}}^{qkv}_j &= \bm{b}^{qkv}_j - (\bm{s}\odot \bm{r}_2) \bm{W}^{qkv}_{:,j}    
\end{aligned}
\end{equation}

Combining Eqs. \ref{eq:raparm_1} and \ref{eq:raparm_2}, the parameters to be transformed are presented in Eq. \ref{eq:raparm}.

\section{More experimental details and results}
\subsection{Implementation Details}
\label{app:exp_details}

\noindent\textbf{Models and datasets}
The pre-trained model we adopt is SAM-Med2D\cite{cheng2023sam}, which achieves impressive image segmentation performance in the medical domain.
To demonstrate generalizability, we adopt a variety of datasets from multiple modalities, as follows:
\begin{itemize}
\item AbdomenCT1k\cite{Ma-2021-AbdomenCT-1K}: Abdominal CT organ segmentation, including the liver, kidneys, spleen, and pancreas. The test dataset includes 3,000 images with 3,361 labels.
\item FLARE\cite{ma2023fast}: Abdominal CT organ segmentation, including 13 regions such as the liver, spleen, pancreas, right kidney, left kidney, and stomach. The test dataset includes 479 images with 2,480 labels.
\item Synapse\cite{synapse}: Abdominal CT organ segmentation, including 9 regions such as the aorta, gallbladder, spleen, left kidney, and right kidney. The test dataset includes 175 images with 672 labels.
\item ACDC\cite{bernard2018deep}: Cardiac MRI organ segmentation, including delineation of the left ventricular cavity, myocardium, and right ventricle. The test dataset includes 423 images with 481 labels.
\item MSD\cite{antonelli2022medical}: MRI organ segmentation, including the heart and prostate. The test dataset includes 190 images with 193 labels.
\item ATLAS\cite{quinton2023tumour}: Liver MRI organ segmentation, including the liver and liver tumors. The test dataset includes 773 images with 1,169 labels.
\item AutoPET\cite{gatidis2022whole}: PET lesion segmentation, including malignant lymphoma, melanoma, and non-small cell lung cancer. The test dataset includes 521 images with 533 labels.
\item ISIC\cite{gutman2016skin}\cite{codella2018skin}: Dermoscopy melanoma segmentation. The test dataset includes 330 images with 330 labels.
\item EndoVis\cite{allan20192017}: Endoscopy abdominal organ segmentation. The test dataset includes 492 images with 648 labels.
\item SA-Med2D-20M\cite{cheng2023sam}: The large medical image dataset collected in SAM-Med 2D, and we use ultrasound, X-ray, and Endoscopy datasets (prefixed with SA). SA-Ultrasound is the segmentation of nerves and breasts, including 198 images with 198 labels; SA-Xray is the chest X-ray segmentation for pneumothorax and pulmonary embolism, including 555 images with 580 labels; SA-Endoscopy is the colonoscopy polyp segmentation, including 126 images with 127 labels.
\end{itemize}

\noindent\textbf{Comparison methods}
First, DFQ-SAM is compared to the full-precision baseline, which is the original pre-trained model without any post-processing. Then, since there is no data-free quantization work for SAM to date, we have to construct the comparison methods on our own under equivalent conditions, as follows:
\begin{itemize}
    \item Real Data: Sampling real data from the original dataset for quantization calibration.
    \item Gaussian Data: Directly using random Gaussian noise for quantization calibration.
\end{itemize}

Note that to ensure a fair comparison, they differ from DFQ-SAM only in calibration data, while remaining identical in other aspects, such as the calibration strategy.

\noindent\textbf{Experimental settings}
All experiments in this work are implemented in Pytorch. To accommodate the hardware and deployment toolchains, we apply channel-wise quantization for weights and layer-wise quantization for activations in the inference process, where scale reparameterization is applied to post-LayerNorm activations in all blocks. The quantization bit precision is set to 4, both for weights and activations.
We synthesize one image for each dataset to perform quantization calibration, with a resolution of 256x256 pixels, which is aligned with the training set of SAM-Med2D. In the data synthesis phase, we adopt the Adam \cite{kingma2014adam} optimizer. Although further tuning might improve accuracy, we standardized the process by setting the total number of iterations to 1500, with label evolution occurring simultaneously during the first 500 iterations. The hyperparameters $\alpha$ and $\beta$ are set to 0.5 and 0.05, respectively, after a grid search. All data synthesis and accuracy evaluation experiments are done on a single NVIDIA RTX A6000 GPU.

\subsection{Visualization of Label Evolution}
\begin{figure*}[t]
\centering
\subfigure[Sample 1, Grayscale Image]{
\includegraphics[width=0.9\linewidth]{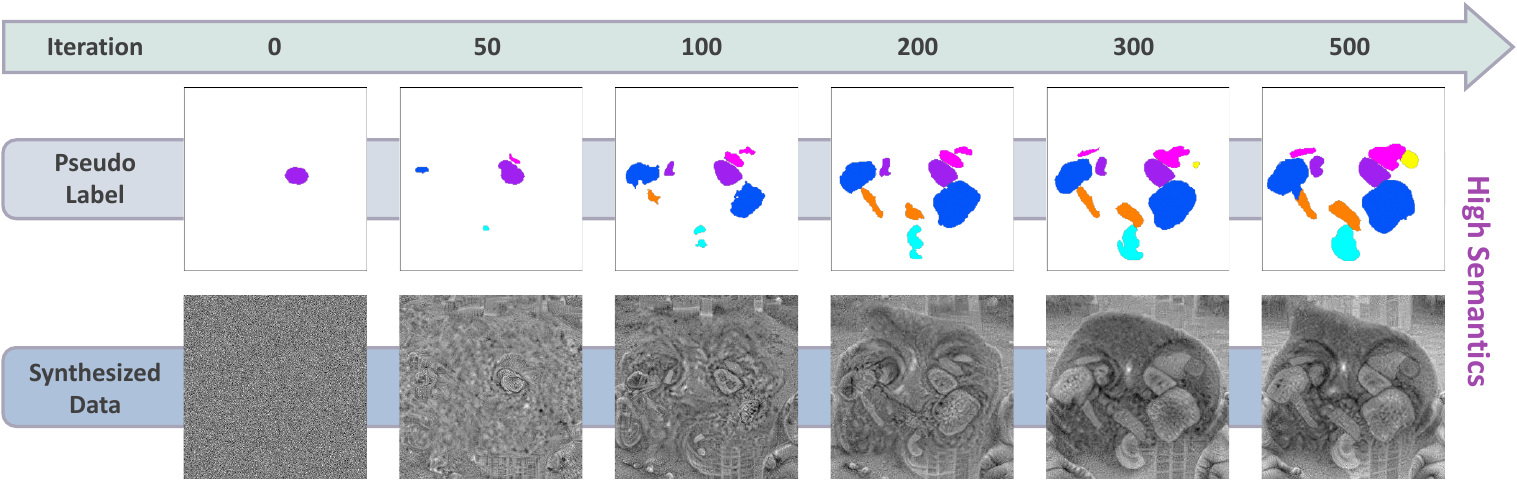}
}\\
\subfigure[Sample 2, Grayscale Image]{
\includegraphics[width=0.9\linewidth]{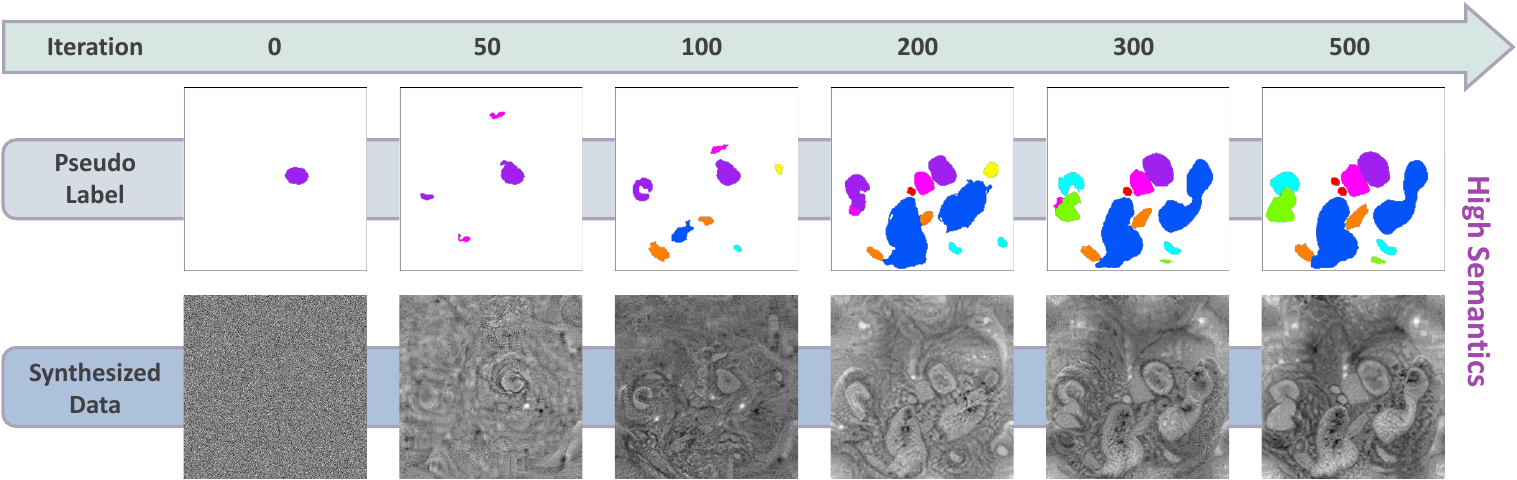}
}\\
\subfigure[Sample 3, Color Image]{
\includegraphics[width=0.9\linewidth]{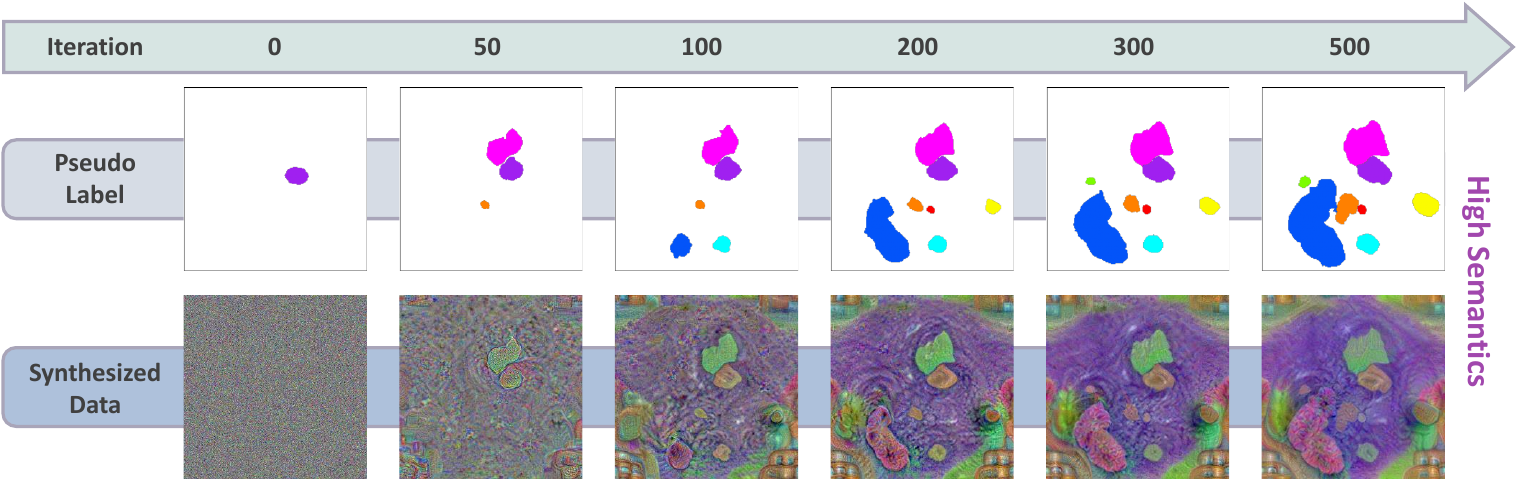}
}\\
\subfigure[Sample 4, Color Image]{
\includegraphics[width=0.9\linewidth]{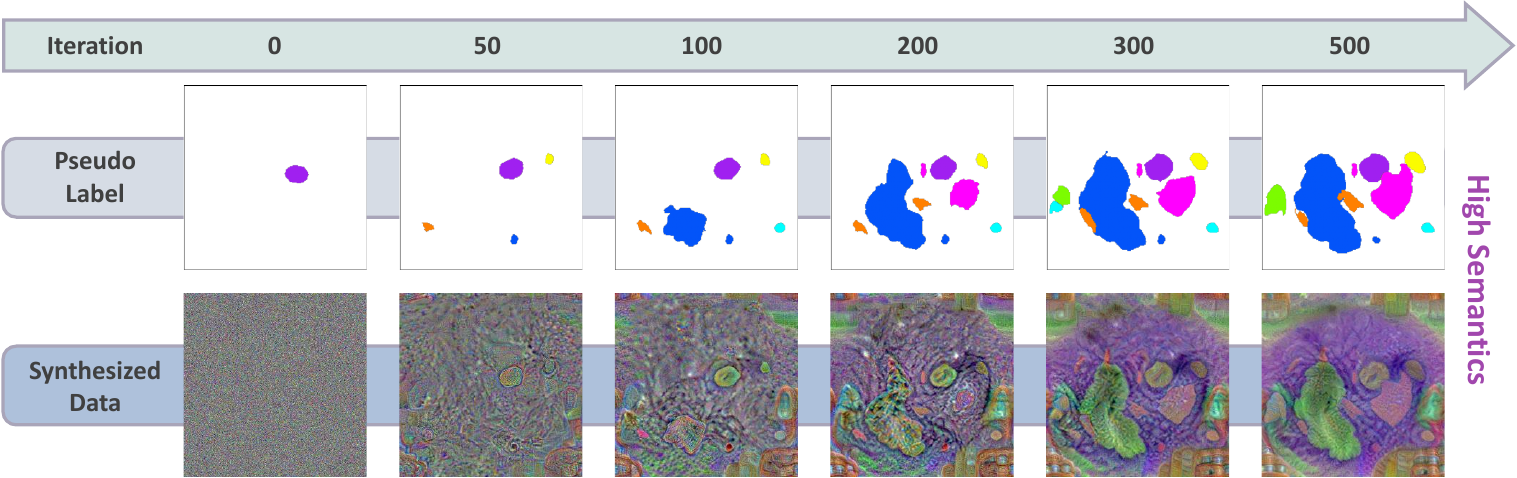}
}
\caption{Illustration of pseudo-label evolution and synthesized image updating (256$\times$256 pixels). Label evolution and image updating are conducted alternately: The label continuously discovers new regions based on the image outputs, which in turn further drives the image optimization. Ultimately, both converge at a high semantic level.}
\label{fig:app_label_vis}
\end{figure*}

Fig \ref{fig:app_label_vis} provides more illustrations of pseudo-label evolution and synthesized image updating (256$\times$256 pixels). In the case of different samples, both labels and images co-evolve and gradually learn more semantic information, resulting in a win-win situation.

\subsection{Qualitative Evaluation of Quantization Results}
\begin{figure*}[t]
\centering
\subfigure[AbdomenCT1k, FLARE, Synapse, and AutoPET Datasets]{
\includegraphics[width=0.95\linewidth]{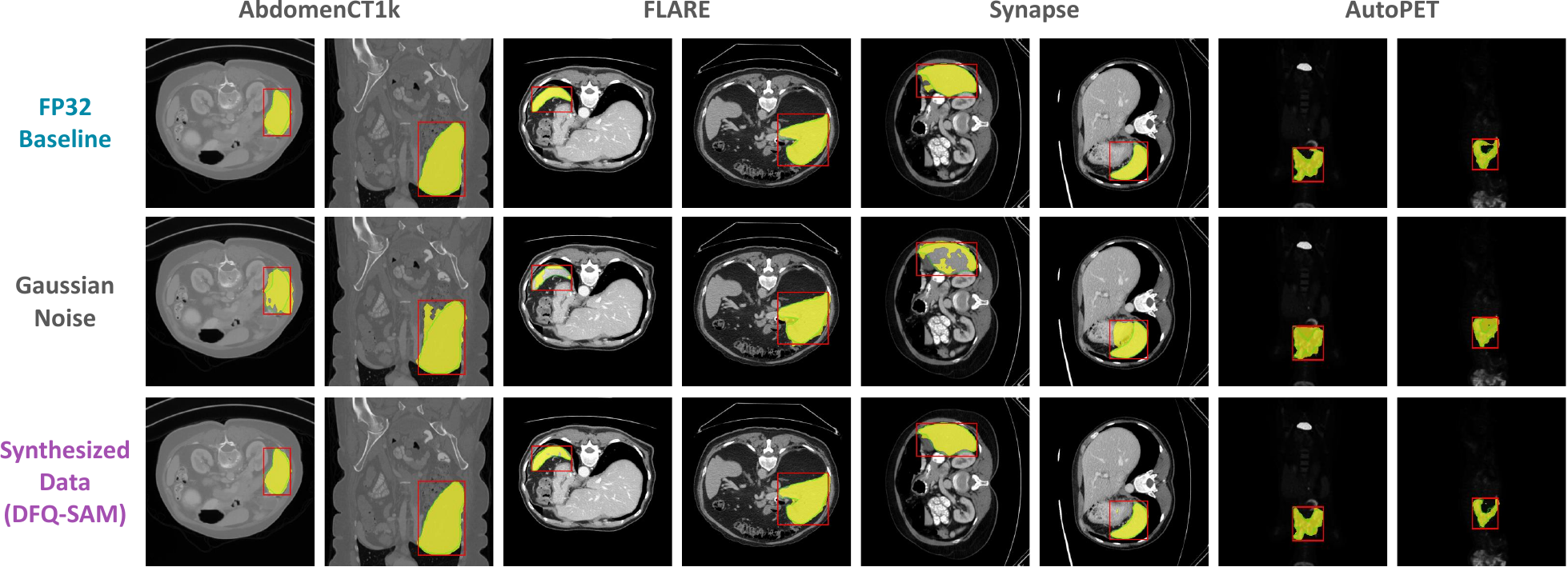}
}\\
\subfigure[ACDC, MSD, ATLAS, and SA-Ultrasound Datasets]{
\includegraphics[width=0.95\linewidth]{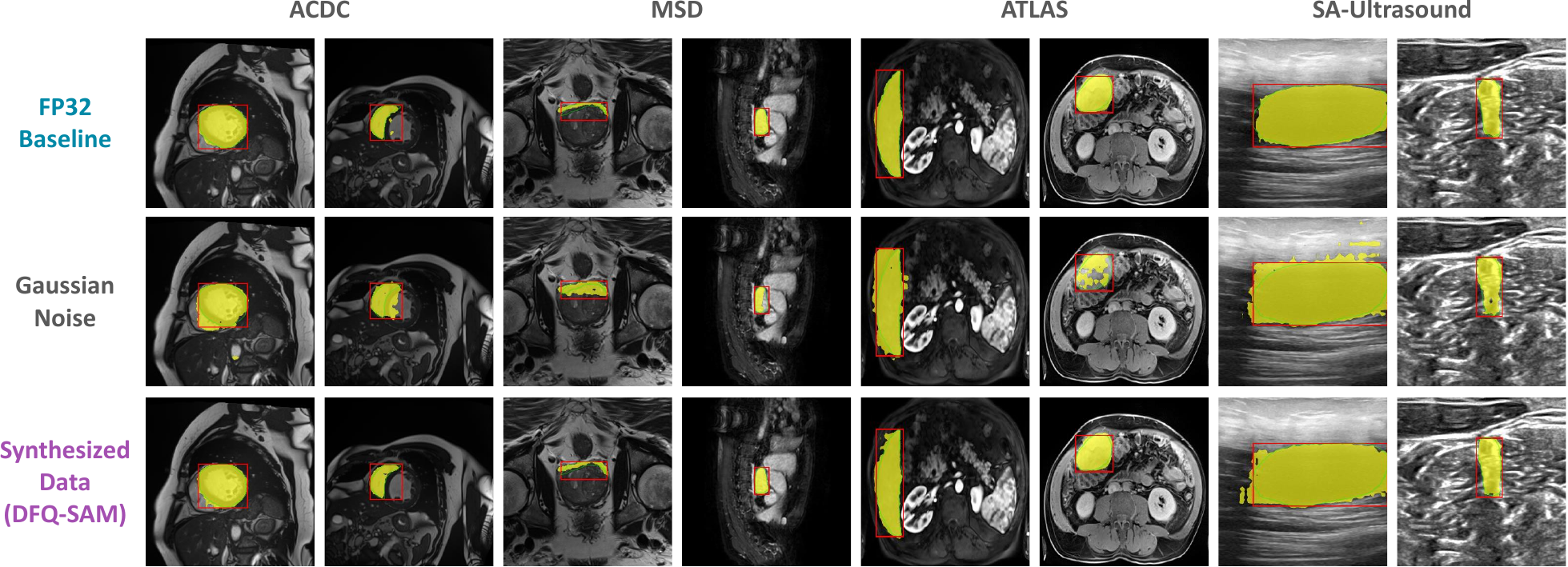}
}\\
\subfigure[SA-Xray, ISIC, EndoVis, and SA-Endoscopy Datasets]{
\includegraphics[width=0.95\linewidth]{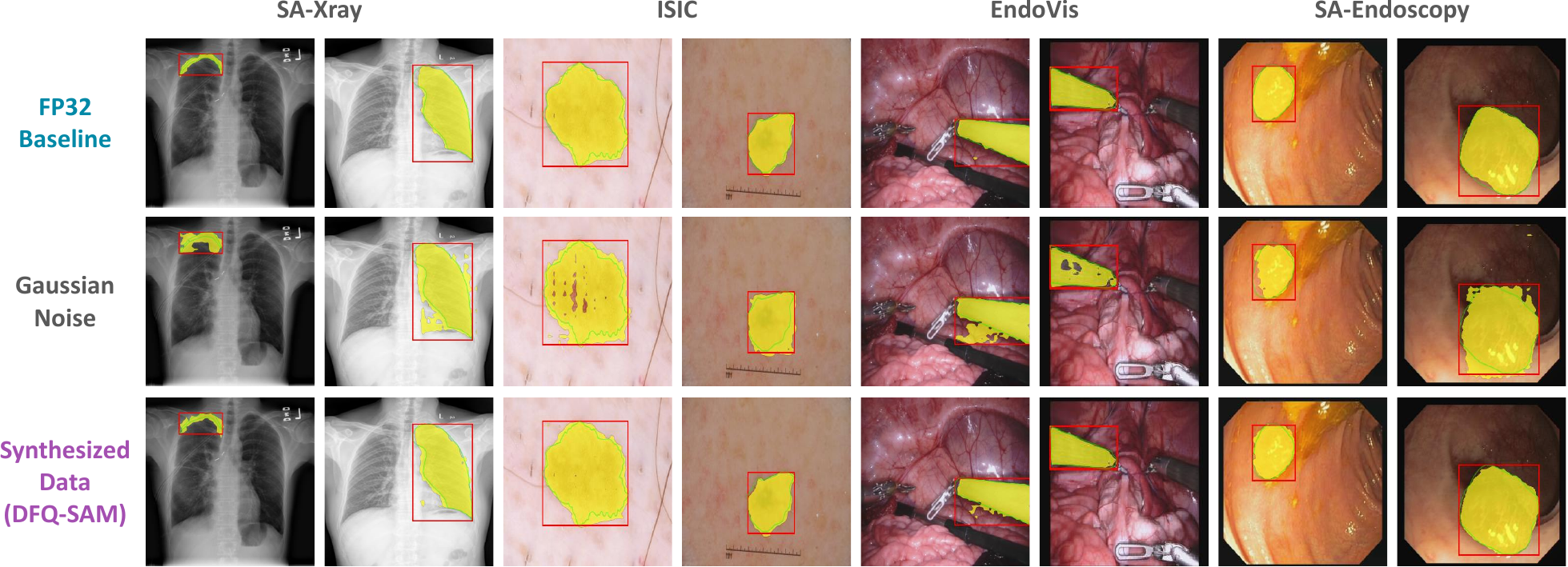}
}
\caption{Visualization of segmentation results on different datasets (256$\times$256 pixels). The proposed DFQ-SAM, which utilizes synthesized data, consistently achieves comparable performance to full-precision baseline at 4-bit quantization.}
\label{fig:app_result_vis}
\end{figure*}

Fig \ref{fig:app_result_vis} provides more visualizations of segmentation results on different datasets (256$\times$256 pixels).
As observed, the synthesized data from our DFQ-SAM offers significantly better quantization calibration than Gaussian noise. Furthermore, the 4-bit quantized model produced by DFQ-SAM achieves performance comparable to the full-precision baseline, enabling almost lossless compression.

%

\end{document}